\documentclass[11pt,a4paper]{article}
\usepackage[hyperref]{emnlp2020}
\usepackage{times}
\usepackage{amsmath}
\usepackage{graphicx}
\usepackage{booktabs}
\usepackage{latexsym}
\usepackage{multirow}
\usepackage{amssymb}
\usepackage{pifont}
\usepackage{microtype}

\aclfinalcopy 
\def\aclpaperid{1581} 

\usepackage[shortlabels]{enumitem}
\setenumerate{itemsep=5pt,topsep=5pt,parsep=0pt,partopsep=5pt,itemindent=5pt}

\title{Reformulating Unsupervised Style Transfer \\ as Paraphrase Generation}

\author{Kalpesh Krishna$^{\spadesuit}$ \And John Wieting$^\diamondsuit$ \\\\ $^\spadesuit$University of Massachusetts Amherst, $^\diamondsuit$Carnegie Mellon University \\ \texttt{\{kalpesh,miyyer\}@cs.umass.edu}\\ \texttt{jwieting@cs.cmu.edu} \\ \\ \textbf{Project Page}: \texttt{\href{http://style.cs.umass.edu}{http://style.cs.umass.edu}} \And Mohit Iyyer$^\spadesuit$}

\date{}

\newcommand{\accuracy}{\textsc{acc}}
\newcommand{\similarity}{\textsc{sim}}
\newcommand{\fluency}{\textsc{fl}}
\newcommand{\dataset}{\textsc{CDS}}
\newcommand{\model}{\textsc{strap}}
\newcommand{\paraEq}{f_\text{para}}
\newcommand{\invParaEq}{f_\text{inv}}

\newcommand{\namedref}[2]{\hyperref[#2]{#1~\ref*{#2}}}

\newcommand{\sectionref}[1]{\namedref{Section}{#1}}
\newcommand{\tableref}[1]{\namedref{Table}{#1}}
\newcommand{\figureref}[1]{\namedref{Figure}{#1}}
\newcommand{\appendixref}[1]{\namedref{Appendix}{#1}}

\begin{document}
\maketitle

\begin{abstract}

Modern NLP defines the task of \emph{style transfer} as modifying the style of a given sentence without appreciably changing its semantics, which implies that the outputs of style transfer systems should be paraphrases of their inputs. However, many existing systems purportedly designed for style transfer inherently warp the  input's meaning through \emph{attribute transfer}, which changes semantic properties such as sentiment. In this paper, we reformulate unsupervised style transfer as a paraphrase generation problem, and present a simple methodology based on fine-tuning pretrained language models on automatically generated paraphrase data. Despite its simplicity, our method significantly outperforms state-of-the-art style transfer systems on both human and automatic evaluations. We also survey 23 style transfer papers and discover that existing automatic metrics can be easily gamed and propose fixed variants. Finally, we pivot to a more real-world style transfer setting by collecting a large dataset of 15M sentences in 11 diverse styles, which we use for an in-depth analysis of our system.

\end{abstract}

\section{Introduction}
\label{sec:introduction}

The task of \emph{style transfer} on text data involves changing the style of a given sentence while preserving its semantics.\footnote{We use the \emph{quasi-paraphrase} definition of semantic equivalence from~\citet{bhagat-hovy-2013-squibs} throughout this paper. We loosely define \emph{style} as patterns in lexical and syntactic choice within the space of quasi-paraphrases.}
Recent work in this area conflates style transfer with the related task of \emph{attribute transfer}~\citep{subramanian2018multiple, he2020probabilistic}, in which modifications to attribute-specific content words (e.g., those that carry sentiment) warp both stylistic \emph{and} semantic properties of a sentence~\citep{preotiuc2016discovering}. Attribute transfer has been criticized for its limited real-world applications:~\citet{pang-2019-towards} argue that semantic preservation is critical for author obfuscation~\citep{shetty2018a4nt}, data augmentation~\citep{xie2019unsupervised, kaushik2020learning}, text simplification~\citep{xu2015problems}, writing assistance~\citep{heidorn2000intelligent}. Moreover, semantic preservation (via paraphrases) has several applications like better translation evaluation~\citep{sellam-etal-2020-bleurt, freitag2020bleu} and adversarial defenses~\citep{iyyer-etal-2018-adversarial}.

We propose to improve  semantic preservation in style transfer by modeling the task as a controlled paraphrase generation problem. Our unsupervised method (\textbf{S}tyle \textbf{Tra}nsfer via \textbf{P}araphrasing, or \model) requires no parallel data between different styles and proceeds in three simple stages:

\begin{figure}[t]
    \centering
    \includegraphics[width=\linewidth]{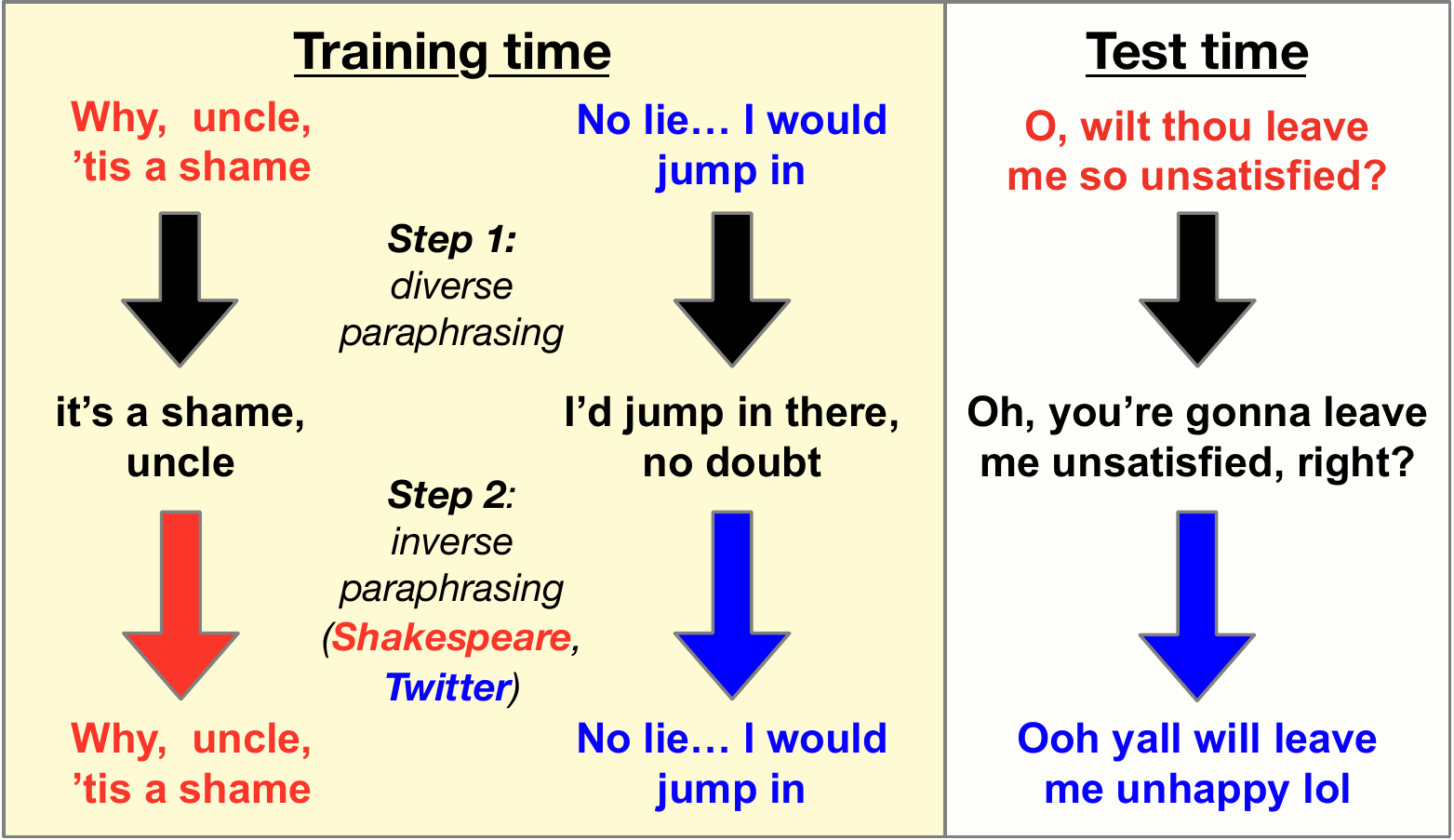}
    \caption{During training, \model\ applies a diverse paraphraser to an input sentence and passes the result through a style-specific \emph{inverse paraphraser} to reconstruct the input. At test time, we perform style transfer by swapping out different inverse paraphrase models (\textcolor{red}{Shakespeare} $\rightarrow$ \textcolor{blue}{Twitter} shown here). All generated sentences shown here are \textbf{actual outputs from \model}.}
    \vspace{-0.15in}
    \label{fig:styletransfer}
\end{figure}

\vspace{-0.1in}
\begin{enumerate}
\setlength\itemsep{0.1em}
    \item Create pseudo-parallel data by feeding sentences from different styles through a diverse paraphrase model (\figureref{fig:styletransfer}, left).
    \item Train style-specific \emph{inverse} paraphrase models that convert these paraphrased sentences back into the original stylized sentences.
    \item Use the inverse paraphraser for a desired style to perform style transfer (\figureref{fig:styletransfer}, right).
\end{enumerate}

Our approach requires none of the finicky\footnote{For example, reproducing deep RL methods is challenging~\citep{henderson2018deep}, vanilla adversarial training is unstable~\citep{arjovsky2017wasserstein}, and VAEs suffer from posterior collapse~\citep{bowman2016generating}. 
} modeling paradigms popular in style transfer research --- no reinforcement learning~\citep{luo2019dual},
variational inference~\citep{he2020probabilistic}, or autoregressive sampling during training~\citep{subramanian2018multiple}. Instead, we implement the first two stages of our pipeline by simply fine-tuning a pretrained GPT-2 language model~\citep{radford2019language}.

Despite its simplicity, \model~significantly outperforms the state of the art on formality transfer and Shakespeare author imitation datasets by \textbf{2-3x} on automatic evaluations and \textbf{4-5x} on human evaluations. We further show that only 3 out of 23 prior style transfer papers properly evaluate their models: in fact, a na\"ive baseline that randomly chooses to either copy its input or retrieve a random sentence written in the target style \textit{outperforms} prior work on poorly-designed metrics.

Finally, we take a step towards real-world style transfer by collecting a large dataset \dataset~(\textbf{C}orpus of \textbf{D}iverse \textbf{S}tyles) of 15M English sentences spanning \textbf{11 diverse styles}, including the works of James Joyce, romantic poetry, tweets, and conversational speech. \dataset~is orders of magnitude larger and more complex than prior benchmarks, which generally focus on transferring between just two styles. We analyze \model's abilities on \dataset, and will release it as a benchmark for future research.

\noindent In summary, \textbf{our contributions} are: 

\noindent\textbf{(1)} a simple approach to perform lexically and syntactically diverse paraphrasing with pretrained language models;

\noindent\textbf{(2)} a simple unsupervised style transfer method that models semantic preservation with our paraphraser and significantly outperforms prior work;

\noindent\textbf{(3)} a critique of existing style transfer evaluation based on a na\"ive baseline that performs on par with prior work on poorly designed metrics;

\noindent\textbf{(4)} a new benchmark dataset that contains 15M sentences from 11 diverse styles.

\section{Style Transfer via Paraphrasing}
\label{sec:approach}

We loosely define \emph{style} as common patterns of lexical choice and syntactic constructions that are distinct from the content of a sentence, following prior work~\citep{hovy1987generating, dimarco1993computational, green1993stylistic, kabbara-cheung-2016-stylistic}. While we acknowledge this distinction is not universally accepted,\footnote{For example, \citet{eckert2008variation} considers style and semantics to be inseparable; while \citet{meyerhoff2015introducing} considers style to be intra-speaker variation in different social contexts} this treatment is critical to unlock several real-world applications of style transfer (as argued in \sectionref{sec:introduction}). Unfortunately, many modern style transfer systems do not respect this definition: a human evaluation (\tableref{tab:prior-human-eval}) shows that \textbf{fewer than 25\% of style-transferred sentences} from two state-of-the-art systems~\citep{subramanian2018multiple, he2020probabilistic} on formality transfer were rated as paraphrases of their inputs.

\begin{figure}[t!]
    \centering
    \includegraphics[width=7.7cm]{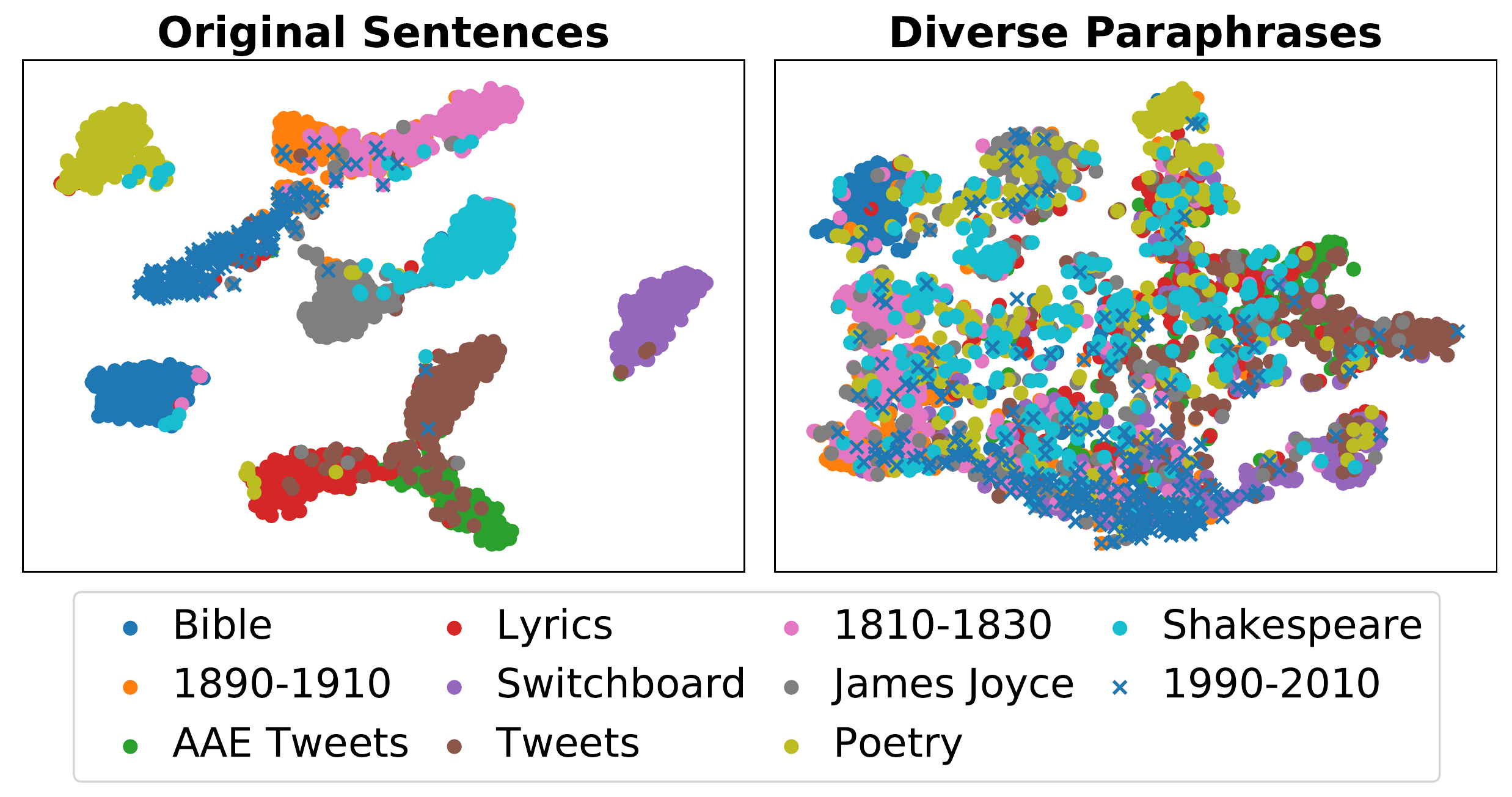}
    \caption{\textbf{Diverse paraphrasing normalizes sentences by removing stylistic identifiers.} We cluster validation sentences from our \dataset\ dataset by applying $t$-SNE to \texttt{[CLS]} vectors from a RoBERTa style classifier. The original sentences (left) form distinct clusters, while the paraphrased sentences (right) do not, showing the stylized text has been normalized.}
    \vspace{-0.1in}
    \label{fig:clusters}
    \vspace{-0.1in}
\end{figure}

Motivated by this result, we reformulate style transfer as a controlled paraphrase generation task. We call our method \textbf{\model}, or \textbf{S}tyle \textbf{Tra}nsfer via \textbf{P}araphrasing. \model~operates within an unsupervised setting: we have raw text from distinct target styles, but no access to parallel sentences paraphrased into different styles. To get around this lack of data, we create \emph{pseudo-parallel} sentence pairs using a paraphrase model (\sectionref{sec:style-normalize}) trained to maximize output diversity (\sectionref{sec:filtering-paraphrase}). Intuitively, this paraphrasing step normalizes the input sentence by stripping away information that is predictive of its original style (\figureref{fig:clusters}). The normalization effect allows us to train an \emph{inverse paraphrase} model specific to the original style, which attempts to generate the original sentence given its normalized version (\sectionref{sec:inverse-paraphrase}). Through this process, the model learns to identify and produce salient features of the original style without unduly warping the input semantics.

\subsection{Creating pseudo-parallel training data}
\label{sec:style-normalize}

The first stage of our approach involves normalizing input sentences by feeding them through a diverse paraphrase model. 
Consider a corpus of sentences from multiple styles,  where the set of all sentences from style $i$ is denoted by $\mathbf{X}^{i}$. We first generate a paraphrase $\mathbf{z}$ for every sentence $\mathbf{x} \in \mathbf{X}^{i}$ using a pretrained paraphrase model $\paraEq$, 
\begin{align*}
    \mathbf{z} &= \paraEq(\mathbf{x}) ~\text{where }\mathbf{x} \in \mathbf{X}^{i}.
\end{align*}

This process results in a dataset $\mathbf{Z}^i$ of normalized sentences and allows us to form a pseudo-parallel corpus $(\mathbf{X}^{i}, \mathbf{Z}^{i})$ between each original sentence and its paraphrased version. \figureref{fig:clusters} shows that this paraphrasing process has a powerful style normalization effect for our instantiation of  $\paraEq$.

\subsection{Style transfer via inverse paraphrasing}
\label{sec:inverse-paraphrase}
We use this pseudo-parallel corpus to train a style-specific model that attempts to reconstruct the original sentence $\mathbf{x}$ given its paraphrase $\mathbf{z}$. Since $\paraEq$ removes style identifiers from its input, the intuition behind this \emph{inverse} paraphrase model is that it learns to insert stylistic features through the reconstruction process. Formally, the inverse paraphrase model $\invParaEq^i$ for style $i$ learns to reconstruct\footnote{This process resembles \textbf{denoising autoencoders}~\citep[][DAE]{vincent2008extracting,lample2017unsupervised}: $\paraEq$ acts as a semantic preserving noise function; $\invParaEq^i$ reconstructs the input.
} the original corpus $\mathbf{X}^i$ using the standard language modeling objective with cross-entropy loss $\mathcal{L}_\text{CE}$,
\begin{align*}
\mathbf{\bar{x}} &= \invParaEq^i(\mathbf{z}) \text{ where } \mathbf{z} \in \mathbf{Z}^i\\
\text{loss} &= \sum_{\mathbf{x} \in \mathbf{X}^i} \mathcal{L}_\text{CE}(\mathbf{x}, \mathbf{\bar{x}}) 
\end{align*}

\noindent During inference, given an arbitrary sentence $\mathbf{s}$ (in any particular style), we convert it to a sentence $\mathbf{\bar{s}}^j$ in target style $j$ using a two-step process of style normalization with  $\paraEq$ followed by stylization with the inverse paraphraser  $\invParaEq^j$, as in
\begin{align*}
\mathbf{\bar{s}}^j &= \invParaEq^j(\paraEq(\mathbf{s})).
\end{align*}

\subsection{Paraphraser implementation with GPT-2}
\label{sec:seq2seq}
We fine-tune the large-scale pretrained GPT2-large language model~\citep{radford2019language} to implement both the paraphraser $\paraEq$ and inverse paraphrasers $\invParaEq^i$ for each style.\footnote{We fine-tune a separate GPT-2 model $\invParaEq^i$ per style. \sectionref{sec:ablation} shows that this outperforms a single inverse paraphraser shared across all styles with style input.} Starting from a pretrained LM improves both output fluency and generalization to small style-specific datasets (\sectionref{sec:ablation}). We use the encoder-free seq2seq modeling approach described in~\citet{wolf2019transfertransfo}, where input and output sequences are concatenated together with a separator token.
We use Hugging Face's Transformers library~\citep{Wolf2019HuggingFacesTS} to implement our models; see \appendixref{appendix:hyperparameter} for more details about the architecture \& hyperparameters.

\subsection{Promoting diversity by filtering data}
\label{sec:filtering-paraphrase}
The final piece to our approach is how we choose training data for the paraphrase model $\paraEq$. We discover that maximizing lexical and syntactic diversity of the output paraphrases is crucial for effective style normalization (\sectionref{sec:ablation}, \ref{sec:analysis}). We promote output diversity by training $\paraEq$ on an aggressively-filtered subset of \textsc{paranmt-50m}~\citep{wieting-gimpel-2018-paranmt}, a large corpus of backtranslated text. Specifically, we apply three filters: (1) removing sentence pairs with more than 50\% trigram or unigram overlap to maximize lexical diversity and discourage copying; (2) removing pairs with lower than 50\% reordering of shared words, measured by Kendall's tau~\citep{kendall1938new}, to promote syntactic diversity; and (3) removing pairs with low semantic similarity, measured by the \textsc{sim} model from \citet{wieting-etal-2019-beyond}.\footnote{This model achieves strong performance on semantic textual similarity (STS) SemEval benchmarks~\citep{agirre-etal-2016-semeval}. We remove all pairs with a score lower than 0.7.} After applying these filters, our training data size shrinks from 50M to 75K sentence pairs, which are used to fine-tune GPT-2; see \appendixref{appendix:paranmt-filtering} for more details about the filtering process and its effect on corpus size.

\section{Evaluating style transfer}
\label{sec:eval_methodology}

Providing a meaningful comparison of our approach to existing style transfer systems is difficult because of (1) poorly-defined automatic and human methods for measuring style transfer quality~\citep{pang-2019-towards, mir-etal-2019-evaluating, tikhonov2019style}, and (2) misleading (or absent) methods of aggregating three individual metrics (transfer accuracy, semantic similarity and fluency) into a single number. In this section, we describe the flaws in existing metrics and their aggregation (the latter illustrated through a na\"ive baseline), and we propose a new evaluation methodology to fix these issues.

\subsection{Current state of style transfer evaluation}

We conduct a survey of \textbf{23} previously-published style transfer papers (more details in \appendixref{appendix:eval-survey}), which reveals three common properties on which style transfer systems are evaluated. Here, we discuss how prior work implements evaluations for each of these properties and propose improved implementations to address some of their downsides.\\

\vspace{-0.1in}

\noindent\textbf{Transfer accuracy (\accuracy):} Given an output sentence $\mathbf{\bar{s}}^{j}$ and a target style $j$, a common way of measuring transfer success is to train a classifier to identify the style of a transferred sentence and report its accuracy \accuracy\ on generated sentences (i.e., whether $\mathbf{\bar{s}}^{j}$ has a predicted style of $j$). 
14 of 23 surveyed papers implement this style classifier with a 1-layer CNN~\citep{kim-2014-convolutional}. However, recent large Transformers like BERT~\citep{devlin-etal-2019-bert} significantly outperform CNNs on most NLP tasks, including style classification. Thus, we build our style classifier by fine-tuning  RoBERTa-large~\citep{liu2019roberta} on all our datasets, leading to significantly more reliable \accuracy\ evaluation.\footnote{The RoBERTa style classifier, built with \texttt{fairseq}~\citep{ott-etal-2019-fairseq}, achieves a test accuracy of 90.4\% on the Shakespeare  data(vs 83.5\% for CNN) and 94.8\% on the Formality data (vs 92.4\%). The datasets are introduced in \sectionref{sec:prior-datasets}.}\\

\vspace{-0.1in}

\noindent\textbf{Semantic similarity (\similarity):} A style transfer system can achieve high \accuracy\ scores without maintaining the semantics of the input sentence, which motivates also measuring how much a transferred sentence deviates in meaning from the input. 
15~/~23 surveyed papers use $n$-gram metrics like BLEU~\citep{papineni-etal-2002-bleu} against reference sentences, often along with self-BLEU with the input, to evaluate semantic similarity. Using BLEU in this way has many problems, including (1) unreliable correlations between $n$-gram overlap and human evaluations of semantic similarity~\citep{callison-burch-etal-2006-evaluating}, (2) discouraging output diversity~\citep{wieting-etal-2019-beyond}, and (3) not upweighting important semantic words over other words~\citep{wieting-etal-2019-beyond, wang2020asking}. These issues motivate us to measure semantic similarity using the subword embedding-based \textsc{sim} model of~\citet{wieting-etal-2019-beyond}, which performs well on semantic textual similarity (STS) benchmarks in SemEval workshops~\citep{agirre-etal-2016-semeval}.\footnote{For reference, we evaluate with BLEU in \appendixref{appendix:more-comparisons}.}\\

\vspace{-0.1in}

\noindent\textbf{Fluency (\fluency):} A system that produces ungrammatical outputs can still achieve high scores on both \accuracy\ and \similarity, motivating a separate measure for fluency. Only 10 out of 23 surveyed papers did a fluency evaluation; 9 of which used language model perplexity, which is a poor measure because (1) it is unbounded and (2) unnatural sentences with common words tend to have low perplexity~\citep{mir-etal-2019-evaluating, pang-2019-towards}. To tackle this we replace perplexity with the accuracy of a RoBERTa-large classifier trained on the CoLA corpus~\citep{warstadt2019neural}, which contains sentences paired with grammatical acceptability judgments. In \tableref{tab:prior-auto-eval}, we show that our classifier marks most reference sentences as fluent, confirming its validity.\footnote{\citet{mir-etal-2019-evaluating} also recommended a similar method to evaluate fluency instead of perplexity, where they train classifiers to distinguish between machine / human sentences.}\\
\vspace{-0.1in}

\noindent\textbf{Human evaluation:}
As automatic evaluations are insufficient for evaluating text generation~\citep{liu-etal-2016-evaluate, novikova-etal-2017-need}, 17 out of 23 surveyed style transfer papers also conduct human evaluation. In our work, we evaluate \similarity~and \fluency~using human evaluations.\footnote{We do not conduct human evaluations for \accuracy~since style classification is difficult for an untrained  crowdsourced worker unfamiliar with the set of target styles.}
As we treat style transfer as a paraphrase generation task, we borrow the three-point scale used previously to evaluate paraphrases~\citep{kok-brockett-2010-hitting, iyyer-etal-2018-adversarial}, which jointly captures \similarity~and \fluency. Given the original sentence and the transferred sentence, annotators on Amazon Mechanical Turk can choose one of three options: \textbf{0} for no paraphrase relationship; \textbf{1} for an ungrammatical paraphrase; and \textbf{2} for a grammatical paraphrase. A total of 150 sentence pairs were annotated per model, with three annotators per pair. More details on our setup, payment \& agreement are provided in \appendixref{appendix:human-evaluation}.

\subsection{Aggregation of Metrics}
\label{sec:joint-evaluation}
So far, we have focused on individual implementations of \accuracy, \similarity, and \fluency. After computing these metrics, it is useful to aggregate them into a single number to compare the overall style transfer quality across systems~\citep{pang-2019-towards}. 
However, only 5 out of the 23 papers aggregate these metrics, either at the corpus level~\citep{xu-etal-2018-unpaired, pang-gimpel-2019-unsupervised} or sentence level~\citep{li-etal-2018-delete}. Even worse, the corpus-level aggregation scheme can be easily gamed. Here, we describe a na\"ive system that outperforms state-of-the-art style transfer systems when evaluated using corpus-level aggregation, and we present a new sentence-level aggregation metric that fixes the issue.\\

\vspace{-0.1in}

\noindent\textbf{The issue with corpus-level aggregation:} Aggregating \accuracy, \similarity, and \fluency\ is inherently difficult because they are inversely correlated with each other~\citep{pang-2019-towards}. Prior work has combined these three scores into a single number using geometric averaging~\citep{xu-etal-2018-unpaired} or learned weights~\citep{pang-gimpel-2019-unsupervised}. However, the aggregation is computed \textit{after} averaging each metric independently across the test set (\textit{corpus-level aggregation}), which is problematic since systems might generate sentences that optimize only a subset of metrics. For example, a Shakespeare style transfer system could output \emph{Wherefore art thou Romeo?} regardless of its input and score high on \accuracy~and \fluency, while a model that always copies its input would score well on \similarity~and \fluency~\citep{pang-2019-towards}.\\

\vspace{-0.1in}

\noindent\textbf{A Na\"ive Style Transfer System:}  To concretely illustrate the problem, we design a na\"ive baseline that exactly copies its input with probability $p$ and chooses a random sentence from the target style corpus for the remaining inputs, where $p$ is tuned on the validation set.\footnote{$p$ = 0.4 / 0.5 for Formality / Shakespeare datasets.}
When evaluated using geometric mean corpus-level aggregation (\textsc{gm} column of \tableref{tab:prior-auto-eval}) this system \emph{outperforms} state of the art methods (\textsc{unmt}, \textsc{dslm}) on the Formality dataset despite not doing any style transfer at all!\\

\vspace{-0.1in}

\noindent\textbf{Proposed Metric:} A good style transfer system should \textbf{jointly} optimize all metrics. The strong performance of the na\"ive baseline with corpus-level aggregation indicates that metrics should be combined at the sentence level \textit{before} averaging them across the test set (\textit{sentence aggregation}). Unfortunately, \textbf{only 3} out of 23 surveyed papers measure absolute performance after \textit{sentence-level aggregation}, and all of them use the setup of~\citet{li-etal-2018-delete}, which is specific to human evaluation with Likert scales. We propose a more general alternative,
\begin{align*}
    J(\accuracy, \similarity, \fluency) = \sum_{x \in \mathbf{X}} \frac{\accuracy(x) \cdot \similarity(x) \cdot \fluency(x)}{|\mathbf{X}|}
\end{align*}

\noindent where $x$ is a sentence from a test corpus $\mathbf{X}$. We treat \accuracy~and \fluency~at a sentence level as a binary judgement, ensuring incorrectly classified or disfluent sentences are automatically assigned a score of 0. As a sanity check, our na\"ive system performs extremely poorly on this new metric (\tableref{tab:prior-auto-eval}), as input copying will almost always yield an \accuracy\  of zero, while random retrieval results in low \similarity.

\begin{table*}[t!]
\small
\begin{center}
\begin{tabular}{ lrrrrr|rrrrr } 
 \toprule
 Model & \multicolumn{5}{c|}{Formality (GYAFC)} & \multicolumn{5}{c}{Shakespeare} \\
 & \accuracy & \similarity & \fluency & \textsc{gm}(\textsc{a,s,f}) & $J$(\textsc{a,s,f}) & \accuracy & \similarity & \fluency & \textsc{gm}(\textsc{a,s,f}) & $J$(\textsc{a,s,f}) \\
 \midrule
\textsc{copy} & 5.2 & 80.1 & 88.4 & 33.3 & 4.2 & 9.6 & 67.1 & 79.1 & 37.1 & 7.2 \\
\textsc{na\"ive} & 58.9 & 38.9 & 89.1 & 58.9 & 7.3 & 49.9 & 34.9 & 78.9 & 51.6 & 4.1\\
\textsc{ref} & 93.3 & 100 & 89.7 & 94.2 & 83.8 & 90.4 & 100 & 79.1 & 89.4 & 70.5 \\
\midrule
\textsc{unmt}~\shortcite{subramanian2018multiple} & \textbf{78.5} & 49.1 & 52.5 & 58.7 & 20.0 & 70.5 & 37.5 & 49.6 & 50.8 & 14.6 \\
\textsc{dlsm}~\shortcite{he2020probabilistic} & 78.0 & 47.7 & 53.7 & 58.5 & 18.6 & 71.1 & 43.5 & 49.4 & 53.5 & 16.3 \\
\midrule
\model~($p = 0.0$) & 67.7 & \textbf{72.5} & \textbf{90.4} & \textbf{76.3} & \textbf{45.5} & 71.7 & \textbf{56.4} & \textbf{85.2} & \textbf{70.1} & \textbf{34.7} \\
\model~($p = 0.6$) & 70.7 & 69.9 & 88.5 & 75.9 & 44.5 & 75.7 & 53.7 & 82.7 & 69.5 & 33.5 \\
\model~($p = 0.9$) & 76.8 & 62.9 & 77.4 & 72.0 & 38.3 & \textbf{79.8} & 47.6 & 71.7 & 64.8 & 27.5 \\
\bottomrule
\end{tabular}
\end{center}
\vspace{-0.1in}
\caption{Automatic evaluation of our method \model~(using different $p$ values for nucleus sampling) against prior state-of-the-art methods (\textsc{unmt}, \textsc{dlsm}), lower bound baselines (\textsc{copy}, \textsc{na\"ive}) and reference sentences (\textsc{ref}). \model~significantly outperforms prior work, especially on our proposed $J(\cdot)$ metric. \textsc{gm} is the geometric mean.}
\vspace{-0.1in}
\label{tab:prior-auto-eval}
\end{table*}

\section{Experiments \& Results}
\label{sec:results}

We evaluate our method (\model) on two existing style transfer datasets, using the evaluation methodology proposed in \sectionref{sec:eval_methodology}. Our system significantly outperforms state of the art methods and the na\"ive baseline discussed in \sectionref{sec:joint-evaluation}.

\subsection{Datasets}
\label{sec:prior-datasets}

We focus exclusively on semantics-preserving style transfer tasks, which means that we do not evaluate on \emph{attribute transfer} datasets such as sentiment, gender, and political transfer. Specifically, we use two standard benchmark datasets for Shakespeare author imitation and formality transfer to compare \model~against prior work.
While both datasets contain parallel data, we only use it to automatically evaluate our model outputs; for training, we follow prior work by using the non-parallel train-validation-test splits from~\citet{he2020probabilistic}.

The \textbf{Shakespeare author imitation} dataset~\citep{xu-etal-2012-paraphrasing} contains 37k training sentences from two styles --- William Shakespeare's original plays,  and their modernized versions. Shakespeare's plays are written in Early Modern English, which has a significantly different lexical (e.g., \textit{thou} instead of \textit{you}) and syntactic distribution compared to modern English. Our second dataset is \textbf{Formality transfer}~\citep{rao-tetreault-2018-dear}, which contains 105k sentences, also from two styles. Sentences are written either in formal or informal modern English. Unlike formal sentences, informal sentences tend to have more misspellings, short forms (\textit{u} instead of \textit{you}), and non-standard usage of punctuation.

\begin{table}[t]
\small
\begin{center}
\begin{tabular}{ llrrrr } 
 \toprule
Dataset & Model & \accuracy & \similarity & $J$(\textsc{a},\textsc{s}) & $J$(\textsc{a},\textsc{s},\textsc{f}) \\ 
\midrule
Form. & \textsc{unmt} & 77.3 & 22.7 & 14.7 & 7.3\\
& \textsc{dlsm} & 78.0 & 24.0 & 15.3 & 10.0\\
\cmidrule{2-6}
& $p = 0.0$ & 71.3 & \textbf{76.0} & \textbf{54.7} & \textbf{41.3} \\
& $p=0.9$ & \textbf{79.3} & 56.7 & 46.0 & 28.0 \\
\midrule
Shak. & \textsc{unmt} & 69.3 & 20.7 & 10.0 & 7.3 \\
& \textsc{dlsm} & 65.3 & 37.3 & 21.3 & 9.3 \\
\cmidrule{2-6}
& $p = 0.0$ & 70.7 & \textbf{79.3} & \textbf{56.0} & \textbf{47.3}\\
& $p=0.9$ & \textbf{74.7} & 54.0 & 38.0 & 24.7\\
\bottomrule
\end{tabular}
\end{center}
\vspace{-0.1in}
\caption{Human evaluation of \model~with greedy decoding ($p=0.0$) and nucleus sampling ($p=0.9$) shows large improvements (\textbf{4-5x}) on both the Formality (Form.) and Shakespeare (Shak.) datasets. Details on metric calculations are provided in \appendixref{appendix:human-evaluation}.}
\vspace{-0.1in}
\label{tab:prior-human-eval}
\end{table}

\subsection{Comparisons against prior work}
\label{sec:compare-prior}
We compare \model~on the Shakespeare / Formality datasets against the following baselines:
\begin{itemize}
    \vspace{-0.05in}
    \setlength\itemsep{0.0em}
    \item \textsc{copy}: a lower bound that simply copies its input, which has been previously used in prior work~\citep{subramanian2018multiple, pang-2019-towards}
    \item \textsc{na\"ive}: our method from \sectionref{sec:joint-evaluation} that randomly either copies its input or retrieves a sentence from the target style
    \item \textsc{ref}: an upper bound computed by evaluating reference sentences using our metrics
    \item \textsc{unmt}: unsupervised neural machine translation from~\citet{subramanian2018multiple}
    \item \textsc{dlsm}: the deep latent sequence model from \citet{he2020probabilistic}, which is currently state-of-the-art on both datasets.\footnote{We use the implementations of both \textsc{unmt} and \textsc{dlsm} made publicly available by~\citet{he2020probabilistic}, and we verify that their \textsc{unmt} model performs on par with reported sentiment transfer numbers  in~\citet{subramanian2018multiple}. The original code of~\citet{subramanian2018multiple} has not been open-sourced.} 
\end{itemize}

\model~significantly outperforms the prior state of the art  (\textsc{dlsm}) on automatic metrics (\tableref{tab:prior-auto-eval}) with a $J(\cdot)$ score of 45.5 (vs 18.6) on Formality and 34.7 (vs 16.3) on Shakespeare. The improvements are even larger when \similarity\ and \fluency\ are measured through human evaluations (\tableref{tab:prior-human-eval}): in this setting, \model~achieves 41.3 (vs 10.0) on Formality and 47.3 (vs 9.3) on Shakespeare. Across the board, \model~significantly improves in \similarity~and \fluency\ while maintaining similar \accuracy. Finally, the large gap between \textsc{ref} and \model~on automatic metrics provides exciting avenues for future research.\footnote{Results with other metrics such as BLEU, as well as comparisons against several other baselines like~\citet{li-etal-2018-delete, prabhumoye-etal-2018-style, luo2019dual, dai-etal-2019-style, sudhakar-etal-2019-transforming} are provided in \appendixref{appendix:more-comparisons}. \model~significantly outperforms all  prior work.}

\section{Ablation studies}
\label{sec:ablation}

In this section, we perform several ablations on \model~to understand which of its components contribute most to its improvements over baselines. Overall, these ablations validate the importance of both paraphrasing and pretraining for style transfer.\\

\vspace{-0.1in}

\noindent \textbf{Paraphrase diversity improves \accuracy:} How critical is diversity in the paraphrase generation step? While our implementation of $\paraEq$ is trained on data that is heavily-filtered to promote diversity, we also build a non-diverse paraphrase model by removing this diversity filtering of \textsc{paraNMT-50M} but keeping all other experimental settings identical. In \tableref{tab:ablation-study}, the \emph{--Div.~PP} rows show a drop in \accuracy\ across both datasets as well as higher \similarity, which in both cases results in a lower $J(\cdot)$ score. A qualitative inspection reveals that the decreased \accuracy\ and increased \similarity\ are both due to a greater degree of input copying, which motivates the importance of diversity.\\

\begin{table}[t!]
\small
\begin{center}
\begin{tabular}{ llrrrr } 
 \toprule
Dataset & Model & \accuracy & \similarity & \fluency & $J$(\textsc{a},\textsc{s},\textsc{f}) \\ 
\midrule
Form. & \model & 67.7 & 72.5 & 90.4 & 45.5 \\
& -- Inf. PP & 27.5 & 78.5 & 88.2 & 20.7 \\
& -- Mult. PP & 63.1 & 72.0 & 90.8 & 42.3 \\
& -- Div. PP & 61.2 & 79.5 & 88.7 & 43.8 \\
& -- GPT2 & 84.6 & 43.8 & 61.7 & 23.1 \\
& GPT2-md & 71.0 & 70.7 & 88.6 & 45.8 \\
& GPT2-sm & 69.1 & 68.6 & 87.6 & 42.9 \\
\midrule
Shak. & \model & 71.7 & 56.4 & 85.2 & 34.7 \\
& -- Inf. PP & 40.1 & 66.1 & 76.3 & 23.3 \\
& -- Mult. PP & 45.9 & 56.5 & 91.1 & 24.8 \\
& -- Div. PP & 49.7 & 64.4 & 82.9 & 28.2 \\
& -- GPT2 & 75.6 & 26.7 & 66.9 & 13.6 \\
& GPT2-md & 73.4 & 54.0 & 86.4 & 34.3 \\
& GPT2-sm & 68.0 & 53.2 & 84.6 & 31.5 \\
\bottomrule
\end{tabular}
\end{center}
\vspace{-0.1in}
\caption{Ablation study using automatic metrics on the Formality (Form.) and Shakespeare (Shak.) datasets.}
\vspace{-0.15in}
\label{tab:ablation-study}
\end{table}

\vspace{-0.1in}

\noindent\textbf{Paraphrasing during inference improves \accuracy:} The diverse paraphraser $\paraEq$ is obviously crucial to train our model, as it creates pseudo-parallel data for training $\invParaEq^i$, but is it necessary during inference? We try directly feeding in the original sentence (without the initial paraphrasing step) to the inverse paraphrase model $\invParaEq^i$ during inference, shown in the \emph{--Inf.~PP} row of \tableref{tab:ablation-study}. While \similarity~and \fluency~are largely unaffected, there is a large drop in \accuracy, bringing down the overall score (45.5 to 20.7 in Formality, 34.7 to 23.3 in Shakespeare). This supports our hypothesis that the paraphrasing step is useful for normalizing the input.\\

\vspace{-0.1in}

\noindent \textbf{LM pretraining is crucial for \similarity\ and \fluency:} As we mainly observe improvements on \fluency\ and \similarity\ compared to prior work, a natural question is how well does \model~perform without large-scale LM pretraining? We run an ablation study by replacing the GPT-2 implementations of $\paraEq$ and $\invParaEq^i$ with LSTM seq2seq models, which are trained with global attention~\citep{luong-etal-2015-effective} using OpenNMT~\citep{klein-etal-2017-opennmt} with mostly default hyperparameters.\footnote{The only hyperparameter we tune is the learning rate schedule. More details in \appendixref{appendix:opennmt}.} As seen in the  \emph{--~GPT2}  row of \tableref{tab:ablation-study}, this model performs competitively with the UNMT / DLSM models on $J$(\accuracy,\similarity,\fluency), which obtain 20.0 / 18.6 on Formality (\tableref{tab:prior-auto-eval}), respectively. However, it is significantly worse than \model, with large drops in \similarity~and \fluency.\footnote{A qualitative inspection of outputs confirms the LSTM struggles to maintain semantics. We suspect this is due to lack of training data ($<$ 75K pairs) to learn a powerful paraphraser.} This result  shows the merit of both our algorithm and the boost that LM pretraining provides.\footnote{Additionally, we note that weaker pretrained language models like GPT2-medium (\emph{GPT2-md}) perform similarly to GPT2-large, while GPT2-small (\emph{GPT2-sm}) is notably worse.}\\

\vspace{-0.1in}

\noindent\textbf{Nucleus sampling trades off \accuracy~for \similarity:} While our best performing system uses a greedy decoding strategy, we experiment with nucleus sampling~\citep{holtzman2020curious} by varying the nucleus $p$ value in both \tableref{tab:prior-auto-eval} and \tableref{tab:prior-human-eval}. As expected, higher $p$ improves diversity and trades off increased \accuracy~for lowered \similarity. We find that $p=0.6$ is similar to greedy decoding on $J(\cdot)$ metrics, but higher $p$ values degrade performance.\\

\vspace{-0.1in}

\noindent\textbf{Multiple inverse paraphrasers perform better than a single style-conditional model:} 
Finally, we explore a more parameter-efficient alternative to training a separate inverse paraphrase model per style. Prior work in conditioned language models feed style-specific tokens as input to a single model to control the output style~\citep{johnson2017google, keskar2019ctrl}. We adopt this strategy by feeding in the average RoBERTa-large \texttt{[CLS]} vectors (by running the pretrained classifier on the training data) as style indicators to a single $\invParaEq^i$. We present results in \tableref{tab:ablation-study} as \emph{--Mult.~PP}. This strategy is not as effective as training a different paraphrase model per style, primarily due to drop in \accuracy. We hypothesize this is due to a conflict between different style distributions in a single model.

\section{Towards Real-World Style Transfer}
\label{sec:analysis}

\begin{table*}[t!]
\footnotesize
\begin{center}
\begin{tabular}{ lp{5.8cm}p{6.8cm} } 
\toprule
Direction & Input & Output \\
\midrule
Tweet $\rightarrow$ Shak. & yall kissing before marriage? & And you kiss'd before your nuptial? \\
Lyrics $\rightarrow$ AAE & It's a good thing you don't have bus fare & It's a good thing u aint gettin no ticket \\
Tweet $\rightarrow$ Poetry & Its so disrespectful I hate itttttt & For 'tis so rude, I have a sick regard, \\
Shak. $\rightarrow$ Tweet & Need you my help? & Are yall okay? Like do you need my help?? \\
AAE $\rightarrow$ Shak. & Aint got nooooo holiday spirit frfr & No spirit of this holiday, gentle man. \\
Poetry $\rightarrow$ Bible & For the foul thief is just at your gate. & the wicked thief art at thy door. \\
1990. $\rightarrow$ Tweet & Now I can't talk to him about it. & I dont wanna talk to him abt that \\
1990. $\rightarrow$ Poetry & I don't know why, but I sensed there was something wrong. & Something felt wrong; I knew not why, \\
Tweet $\rightarrow$ Bible & when ur going thru an episode and cant cope anymore & when thou art in an episode, and cannot stand it any more;\\
1890. $\rightarrow$ AAE & I was just thinking the same thing that you have expressed. & u said the same thing i thought lol \\
1990. $\rightarrow$ 1810. & He was being terrorized into making a statement by the same means as the other so-called ``witnesses.''& Terror had been employed in the same manner with the other witnesses, to compel him to make a declaration. \\
AAE $\rightarrow$ Shak. & If I got a dollar every time one of my friends told me they hate me, I'd be rich & I would have been rich, had I but a dollar for every friend that hath said they hate me. \\
Joyce $\rightarrow$ Bible & I appeal for clemency in the name of the most sacred word our vocal organs have ever been called upon & I beseech thee in the name of the most holy word which is in our lips, forgive us our trespasses. \\
\bottomrule
\end{tabular}
\end{center}
\vspace{-0.1in}
\caption{Example outputs from \model~trained on our \dataset~dataset (more generations in \appendixref{appendix:example-generations}).}
\vspace{-0.1in}
\label{tab:generations-1}
\end{table*}

\begin{table}[t]
\small
\begin{center}
\begin{tabular}{ lr|lr } 
 \toprule
 Style & Size & Style & Size \\
 \midrule
 Shakespeare & 27.5K & Lyrics & 5.1M\\
 James Joyce & 41.2K & 1810-1830 & 216.0K\\
 English Tweets & 5.2M & 1890-1910 & 1.3M\\
 AAE Tweets & 732.3K & 1990-2010 & 2.0M\\
 Romantic Poetry & 29.8K & Bible & 34.8K \\
 Switchboard & 148.8K\\
\bottomrule
\end{tabular}
\end{center}
\vspace{-0.1in}
\caption{List of styles in our dataset along with the their total sizes. The year periods (like ``1810-1830'') refer to sentences from the Corpus of Historical American English~\citep{davies2012expanding}. ``AAE Tweets'' refers to African American English Tweets corpus from~\citet{blodgett-etal-2016-demographic}. ``Switchboard'' is a collection of conversational speech transcripts from~\citet{godfrey1992switchboard}. Details of the collection and examples are in \appendixref{appendix:dataset}.}
\vspace{-0.1in}
\label{tab:dataset-mini-stats}
\end{table}

\begin{figure}[t]
    \includegraphics[width=6.5cm]{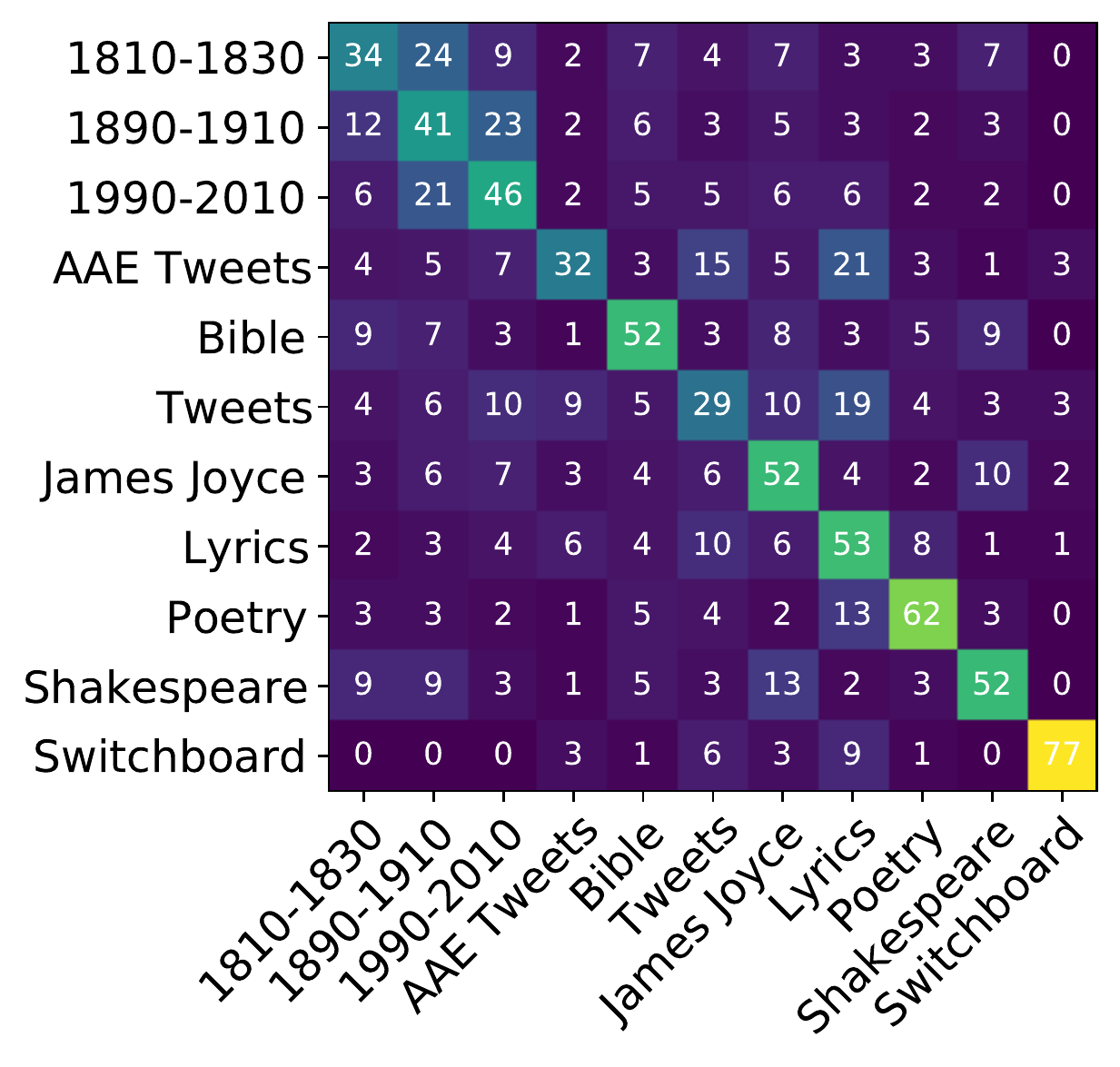}
    \vspace{-0.1in}
    \caption{\textbf{Classifier confusion after style transfer}. Every row shows the classifier label distribution on sentences \emph{transferred to} the target style (the row label). The off-diagonal elements show mis-classifications with intuitive domain similarities, such as (Lyrics, Poetry); (AAE, Tweets); (Joyce, Shakespeare).
    }
    \vspace{-0.1in}
    \label{fig:heatmaps}
    \vspace{-0.1in}
\end{figure}

\begin{table}[t]
\small
\begin{center}
\begin{tabular}{ lrrrr } 
 \toprule
 \multicolumn{5}{c}{Shakespeare $\leftrightarrow$ English Tweets, \dataset} \\
 \midrule
Model & \accuracy & \similarity & \fluency & $J$(\textsc{a},\textsc{s},\textsc{f}) \\ 
\midrule
\textsc{copy} & 0.1 & 100.0 & 69.2 & 0.0 \\
\textsc{unmt}~\shortcite{subramanian2018multiple} & 76.7 & 20.6 & 37.7 & 4.4 \\
\textsc{dlsm}~\shortcite{he2020probabilistic} & 64.2 & 19.6 & 33.1 & 2.0 \\
\midrule
\model~($p = 0.0$) & 20.3 & 65.0 & 81.1 & 8.7 \\
\model~($p = 0.6$) & 31.1 & 58.1 & 75.0 & 10.8 \\
\model~($p = 0.9$) & 43.2 & 54.5 & 68.3 & \textbf{13.9} \\
\bottomrule
\end{tabular}
\end{center}
\vspace{-0.1in}
\caption{A controlled comparison between models on 2 styles from \dataset~using automatic evaluation. \accuracy~is calculated using our \textbf{11-way} \dataset~classifier and \similarity~is with input. \model~greatly outperforms prior work.}
\vspace{-0.1in}
\label{tab:controlled-cds-eval}
\end{table}

All of our experiments and ablations thus far have been on the Shakespeare and Formality datasets, which contain just two styles each.  To explore the ability of our system to perform style transfer between many diverse styles, we create the \textbf{C}orpus of \textbf{D}iverse \textbf{S}tyles (\dataset), a new non-parallel style transfer benchmark dataset with 11 diverse styles (15M tokens), and use it to evaluate \model.\\

\vspace{-0.1in}

\noindent\textbf{Corpus of Diverse Styles}: To create \dataset, we obtain data  (\tableref{tab:dataset-mini-stats}) from existing academic research datasets~\citep{godfrey1992switchboard, blodgett-etal-2016-demographic} and public APIs or online collections like Project Gutenberg~\citep{hart1992history}. We choose styles that are easy for human readers to identify at a sentence level (e.g., Tweets or Biblical text), and the left side of~\figureref{fig:clusters} confirms that machines also cluster \dataset\ into eleven distinct styles.  While prior benchmarks involve a transfer between two styles, \dataset\ has 110 potential transfer directions. We present dataset examples, details on collection and style similarity analysis in \appendixref{appendix:dataset}.\\

\vspace{-0.1in}

\noindent\textbf{Diverse paraphrasing normalizes stylized text}
\label{sec:diverse-paraphrase-effect}
With eleven styles, we can better validate the effectiveness of our diverse paraphraser at normalizing input sentences. After training an 11-way style classifier on \dataset\ using RoBERTa-large,  we observe an accuracy of \textbf{88.9\%} on the original validation set. After paraphrasing the validation set with $\paraEq$, this classifier only correctly classifies \textbf{42.5\%} sentences, indicating a significant decrease in recognizable stylistic features. \figureref{fig:clusters} further demonstrates this normalization effect. Finally, the magnitude of normalization is lower with the non-diverse paraphraser (from \sectionref{sec:ablation}), with a smaller accuracy drop to \textbf{51.5\%} after paraphrasing;\footnote{Even if we retrain the classifiers on a paraphrased version of the training set (to model the distribution better), the performance is only 65.8\% for the diverse model and 72.3\% for the non-diverse model, indicating a loss in style signal.} qualitatively, the diverse model exhibits more lexical swaps and syntactic diversity.\footnote{On average, the diverse model has 51\% unigram F1 word overlap and 27\% word shuffling measured by Kendall's $\tau_\text{B}$, compared to 28\% unigram F1 and 6\% shuffling for the non-diverse model; \appendixref{appendix:paraphrase-our-dataset} has a style-wise breakdown.}\\

\vspace{-0.1in}

\begin{table*}[t!]
\footnotesize
\begin{center}
\begin{tabular}{ lp{7.3cm}p{5.3cm} } 
\toprule
Direction & Input $\rightarrow$ Paraphrase $\rightarrow$ Output & Analysis \\
\midrule
Shak. $\rightarrow$ Bible & Have you importuned him by any means? $\rightarrow$ \newline did you ever try to import him? $\rightarrow$ \newline hast thou ever tried to import him? & Misunderstanding the word ``importune'' --- the model believes it refers to \textit{import} rather than \textit{harass / bother}. \\
\midrule
1990. $\rightarrow$ Tweet. & The machine itself is made of little straws of carbon. $\rightarrow$ \newline the machine is made of straw. $\rightarrow$ \newline Machine made of straw. & Dropping of important semantic words during diverse paraphrasing (``carbon'') significantly warps the meaning of sentences\\
\midrule
Swit. $\rightarrow$ Shak. & well they offer classes out at uh Ray Hubbard $\rightarrow$ \newline they're offering a course at Ray Hubbard's. $\rightarrow$ \newline They do offer a course at the house of the Dukedom. & Hallucination of tokens irrelevant to the input (``house of the dukedom'') to better reflect style distribution.\\
\midrule
Tweet $\rightarrow$ Swit. & Knoxville aint for me $\rightarrow$ I'm not in Knoxville. $\rightarrow$ \newline i don't know Knoxville & Subtle modifications in semantics since the models fail to understand their inputs.\\
\bottomrule
\end{tabular}
\end{center}
\vspace{-0.1in}
\caption{Representative examples showing the common failure modes of \model~when evaluated on \dataset.}
\vspace{-0.1in}
\label{tab:qualitative}
\end{table*}

\noindent\textbf{Style Transfer on \dataset}: We measure \model's performance on \dataset\ using \sectionref{sec:eval_methodology}'s evaluation methodology. We sample 1K sentences from each style and use \model\ to transfer these sentences to each of the 10 other styles.
Despite having to deal with many more styles than before, our system achieves \textbf{48.4\%} transfer accuracy (on a 11-way RoBERTa-large classifier), a paraphrase similarity score of \textbf{63.5}, and \textbf{71.1\%} fluent generations, yielding a $J$(\accuracy,\similarity,\fluency) score of 20.7. A breakdown of style-specific performance is provided in \appendixref{appendix:performance-our-dataset}. An error analysis shows that the classifier misclassifies some generations as styles sharing properties with the target style (\figureref{fig:heatmaps}).\\

\vspace{-0.1in}

\noindent\textbf{Controlled comparisons}: To ground our \dataset~results in prior work, we compare \model~with baselines from \sectionref{sec:compare-prior}. We sample equal number of training sentences from two challenging styles in \dataset~(Shakespeare, English Tweets) and train all three models (\textsc{unmt}, \textsc{dlsm}, \model) on this subset of \dataset.\footnote{We could not find an easy way to perform 11-way style transfer in the baseline models without significantly modifying their codebase / model due to the complex probabilistic formulation beyond 2 styles and separate modeling for each of the 110 directions.} As seen in \tableref{tab:controlled-cds-eval}, \model~greatly outperforms prior work, especially in \similarity~and \fluency. Qualitative inspection shows that baseline models often output arbitrary style-specific features, completely ignoring input semantics (explaining poor \similarity~but high \accuracy).\\

\vspace{-0.1in}

\noindent \textbf{Qualitative Examples}: \tableref{tab:generations-1} contains several outputs from \model; see \appendixref{appendix:example-generations} for more examples. We also add more qualitative analysis of the common failures of our system in \tableref{tab:qualitative}. Our model makes mistakes similar to contemporary text generation systems --- poor understanding of rare words, dropping / modification of semantic content, hallucination to better reflect training distribution.

\section{Related Work}

Unsupervised style transfer is often modeled by disentangling style \& content using attribute classifiers~\citep{hu2017toward, shen2017style}, policy gradient training~\citep{xu-etal-2018-unpaired, luo2019dual} or retrieval-based approaches~\citep{li-etal-2018-delete}. Recently, backtranslation has emerged as a method to model semantic preservation~\citep{prabhumoye-etal-2018-style}, but this method can also warp semantics as seen in~\citet{subramanian2018multiple}; as such, we only use it to build our paraphraser's training data after heavy filtering. Our work relates to recent efforts that use Transformers in style transfer~\citep{sudhakar-etal-2019-transforming,dai-etal-2019-style}. 
Closely related to our work is \citet{grondahl2019effective}, who over-generate paraphrases using a complex handcrafted pipeline and filter them using proximity to a target style corpus. Instead, we \emph{automatically learn} style-specific paraphrasers and do not need over-generation at inference. Relatedly,~\citet{preotiuc2016discovering} present qualitative style transfer results with statistical MT paraphrasers. Other, less closely related work on control \& diversity in text generation is discussed in \appendixref{appendix:more_related}.

\section{Conclusion}

In this work we model style transfer as a controlled paraphrase generation task and present a simple unsupervised style transfer method using diverse paraphrasing. We critique current style transfer evaluation using a survey of 23 papers and propose fixes to common shortcomings. Finally, we collect a new dataset containing 15M sentences from 11 diverse styles. Possible future work includes (1) exploring other applications of diverse paraphrasing, such as data augmentation; (2) performing style transfer at a paragraph level; (3) performing style transfer for styles unseen during training, using few exemplars provided during inference.
\section*{Acknowledgements}

We thank the anonymous reviewers and area chair for their useful comments. We are grateful to Su Lin Blodgett, Katie Keith, Brendan O'Connor, Tu Vu, Shufan Wang, Nader Akoury, Rajarshi Das, Andrew Drozdov, Rico Angell, and the rest of the UMass NLP group for help at various stages of the project. Finally, we thank Arka Sadhu, Richard Pang, Kevin Gimpel, Graham Neubig, Diyi Yang, Shrimai Prabhumoye, Junxian He and Yonatan Belinkov for insightful discussions. This research was supported in part by a research gift from Adobe.

\bibliographystyle{acl_natbib}
\bibliography{bib/journal-full,bib/emnlp2020}
\newpage
\appendix

\section{Appendices for ``Reformulating Unsupervised Style Transfer as Paraphrase Generation''}
\label{sec:appendix}

\subsection{\textsc{paraNMT-50M} Filtering Details}
\label{appendix:paranmt-filtering}

We train our paraphrase model in a seq2seq fashion using the \textsc{paraNMT-50M} corpus~\citep{wieting-gimpel-2018-paranmt}, which was constructed by backtranslating~\citep{sennrich-etal-2016-improving} the Czech side of the CzEng parallel corpus~\citep{czeng16:2016}. This corpus is large and noisy and we aggressively filter it to encourage content preservation and diversity maximization. We use the following filtering,

\noindent \textbf{Content Filtering}: We remove all sentence pairs which score lower than 0.5 on a strong paraphrase similarity model from~\citet{wieting-etal-2019-beyond}.\footnote{We use the \textsc{sim} model from~\citet{wieting-etal-2019-beyond}, which achieves a strong performance on the SemEval semantic text similarity (STS) benchmarks~\citep{agirre-etal-2016-semeval}} We filter sentence pairs by length, allowing a maximum length difference of 5 words between paired sentences. Finally, we remove very short and long sentences by only keeping sentence pairs with an average token length between 7 and 25.\\
\noindent \textbf{Lexical Diversity Filtering}: We only preserve backtranslated pairs with sufficient unigram distribution difference. We filter all pairs where more than 50\% of words in the backtranslated sentence can be found in the source sentence. This is computed using the SQuAD evaluation scripts~\citep{rajpurkar-etal-2016-squad}. Additionally, we remove sentences with more than 70\% trigram overlap.\\
\noindent \textbf{Syntactic Diversity Filtering}: We discard all paraphrases which have a similar word ordering. We compare the relative ordering of the words shared between the input and backtranslated sentence by measuring the Kendall tau distance~\citep{kendall1938new} or the ``bubble-sort'' distance. We keep all backtranslated pairs which are at least 50\% shuffled.\footnote{An identical ordering of words is 0\% shuffled whereas a reverse ordering is 100\% shuffled.}\\
\noindent \textbf{LangID Filtering:} Finally, we discard all sentences where both the input and backtranslated sentence are classified as non-English using \texttt{langdetect}.\footnote{This is using the Python port of ~\citet{nakatani2010langdetect}, \url{https://github.com/Mimino666/langdetect}.}
\\
\noindent \textbf{Effect of each filter}: We adopt a pipelined approach to filtering. The \textsc{paraNMT-50M} corpus size after each stage of filtering is shown in \tableref{tab:filtering-steps}.

\begin{table}[t]
\begin{center}
\begin{tabular}{ llr } 
 \toprule
& Filter Stage & Corpus Size \\ 
\midrule
0. & Original & 51.41M \\
1. & Content Similarity & 30.49M \\
2. & Trigram Diversity & 9.03M \\
3. & Unigram Diversity & 1.96M \\
4. & Kendall-Tau Diversity & 112.01K \\
5. & Length Difference & 82.64K \\
6. & LangID & 74.55K \\
\bottomrule
\end{tabular}
\end{center}
\vspace{-0.1in}
\caption{Steps of filtering conducted on \textsc{paraNMT-50M} along with its effect on corpus size.}
\vspace{-0.15in}
\label{tab:filtering-steps}
\end{table}

\subsection{Generative Model Details}
\label{appendix:hyperparameter}

This section provides details of our seq2seq model used for both paraphrase model and style-specific inverse paraphrase model. Recent work~\cite{radford2019language} has shown that GPT2, a massive transformer trained on a large corpus of unlabeled text using the language modeling objective, is very effective in performing more human-like text generation. We leverage the publicly available GPT2-large checkpoints by finetuning it on our custom datasets with a small learning rate. However, GPT2 is an unconditional language model having only a decoder network, and traditional seq2seq setups use separate encoder and decoder neural network~\citep{sutskever2014sequence} with attention~\citep{bahdanau2014neural}. To avoid training an encoder network from scratch, we use the encoder-free seq2seq modeling approach described in~\citet{wolf2019transfertransfo}.  where both input and output sequences are fed to the decoder network separated with a special token, and use separate segment embeddings. Our model is implemented using the \texttt{transformers} library\footnote{\url{https://github.com/huggingface/transformers}}~\citep{Wolf2019HuggingFacesTS}. We use encoder-free seq2seq modeling~\citep{wolf2019transfertransfo} which feeds the input into the decoder neural network, separating it with segment embeddings. We fine-tune GPT2-large to perform encoder-free seq2seq modeling.\\

\noindent\textbf{Architecture:} Let $\mathbf{x} = (x_1, ..., x_n)$ represent the tokens in the input sequence and let $\mathbf{y} = (y_{bos}, y_1, ..., y_m, y_{eos})$ represent the tokens of the output sequence, where $y_{bos}$ and $y_{eos}$ corresponds to special beginning and end of sentence tokens. We feed the sequence $(x_1 ,..., x_n, y_{bos}, y_1 ,..., y_m)$ as input to GPT2 and train it on the next-word prediction objective for the tokens $y_1,...,y_m, y_{eos}$ using the cross-entropy loss. During inference, the sequence $(x_1 ,..., x_n, y_{bos})$ is fed as input and the tokens are generated in an autoregressive manner~\citep{vaswani2017attention} until $y_{eos}$ is generated.

Every token in $\mathbf{x}$ and $\mathbf{y}$ is passed through a shared input embedding layer to obtain a vector representation of every token. To encode positional and segment information, learnable positional and segment embeddings are added to the input embedding consistent with the GPT2 architecture. Segment embeddings are used to denote whether a token belongs to sequence $\mathbf{x}$ or $\mathbf{y}$.\\

\noindent\textbf{Other seq2seq alternatives:} Note that our unsupervised style transfer algorithm is agnostic to the specific choice of seq2seq modeling. We wanted to perform transfer learning from massive left-to-right language models like GPT2, and found the encoder-free seq2seq approach simple and effective. Future work includes finetuning more recent models like T5~\citep{raffel2019exploring} or BART~\citep{lewis2019bart}. These models use the standard seq2seq setup of separate encoder / decoder networks and pretrain them jointly using denoising autoencoding objectives based on language modeling.\\

\noindent\textbf{Hyperparameter Details:} We finetune GPT2-large using NVIDIA TESLA M40 GPUs for 2 epochs using early stopping based on validation set perplexity. The models are finetuned using a small learning rate of 5e-5 and converge to a good solution fairly quickly as noticed by recent work~\citep{li2020train, kaplan2020scaling}. Specifically, each experiment completed within a day of training on a single GPU, and many experiments with small datasets took a lot less time. We use a minibatch size of 10 sentence pairs and truncate sequences which are longer than 50 subwords in the input or output space. We use the Adam optimizer~\citep{kingma2015adam} with the weight decay fix and using a linear learning rate decay schedule, as implemented in the \texttt{transformers} library. Finally, we left-pad the input sequence to get a total input length of 50 subwords and right-pad output sequence to get a total output length of 50 subwords. This special batching is necessary to use minibatches during inference time. Special symbols are used to pad the sequences and they are not considered in the cross-entropy loss. Our model has 774M trainable parameters, identical to the original GPT2-large.

\subsection{Classifier Model Details}
\label{appendix:classifier-hyperparameters}

We fine-tune RoBERTa-large to build our classifier, using the official implementation in \texttt{fairseq}. We use a learning rate of 1e-5 for all experiments with a minibatch size of 32. All models were trained on a single NVIDIA RTX 2080ti GPU, with gradient accumulation to allow larger batch sizes. We train models for 10 epochs and use early stopping on the validation split accuracy. We use the Adam optimizer~\citep{kingma2015adam} with modifications suggested in the RoBERTa paper~\citep{liu2019roberta}. Consistent with the suggested hyperparameters, we use a learning rate warm-up for the first 6\% of the updates and then decay the learning rate.

\subsection{OpenNMT Model Details}
\label{appendix:opennmt}

We train sequence-to-sequence models with attention based on LSTMs using OpenNMT~\citep{klein-etal-2017-opennmt} using their PyTorch port.\footnote{\url{https://github.com/OpenNMT/OpenNMT-py}} We mostly used the default hyperparameter settings of \texttt{OpenNMT-py}. The only hyperparameter we modified was the learning rate schedule, since our datasets were small and overfit quickly. For the paraphrase model, we started decay after 11000 steps and halved the learning rate every 1000 steps. For Shakespeare, we started the decay after 3000 steps and halved the learning rate every 500 steps. For Formality, we started the decay after 6000 steps and halved the learning rate every 1000 steps. These modifications only slightly improved validation perplexity (by 3-4 points in each case).

We used early stopping on validation perplexity and checkpoint the model every 500 optimization steps. The other hyperparameters are the default \texttt{OpenNMT-py} settings --- SGD optimization using learning rate 1.0, LSTM seq2seq model  with global attention~\citep{luong-etal-2015-effective}, 500 hidden units and embedding dimensions and 2 layers each in the encoder and decoder.

\subsection{More Comparisons with Prior Work}
\label{appendix:more-comparisons}

Please refer to \tableref{tab:prior-auto-eval-bleu} for an equivalent of \tableref{tab:prior-auto-eval} using BLEU scores.

We present more comparisons with prior work in \tableref{tab:prior-auto-eval-luo}. We use the generated outputs for the Formality test set available in the public repository of~\citet{luo2019dual} (including outputs from the algorithms described in~\citealp{prabhumoye-etal-2018-style} and ~\citealp{li-etal-2018-delete}) and run them on our evaluation pipeline. We compare the results with our formality transfer model used in \tableref{tab:prior-auto-eval} and \tableref{tab:prior-human-eval}. We note significant performance improvements, especially in the fluency of the generated text. Note that there is a domain shift for our model, since we trained our model using the splits of~\citet{he2020probabilistic} which use the Entertainment \& Music splits of the Formality corpus. The outputs in the repository of~\citet{luo2019dual} use the Family \& Relationships split. It is unclear in the paper of~\citet{luo2019dual} whether the models were trained on the Family \& Relationships training split or not.\\

\noindent\textbf{Other Comparisons:} We tried to compare against other recent work in style transfer based on Transformers, such as \citet{dai-etal-2019-style} and \citet{sudhakar-etal-2019-transforming}. Both papers do not evaluate their models on the datasets we use (Shakespeare and Formality), where parallel sentences preserve semantics.

The only datasets used in~\citet{dai-etal-2019-style} were sentiment transfer benchmarks, which modify semantic properties of the sentence. We attempted to train the models in~\citet{dai-etal-2019-style} using their codebase on the Shakespeare dataset, but faced three major issues 1) missing number of epochs / iterations. The early stopping criteria is not implemented or specified, and metrics were being computed on the \textit{test set} every 25 training iterations, which is invalid practice for choosing the optimal checkpoint; 2) specificity of the codebase to the Yelp sentiment transfer dataset in terms of maximum sequence length and evaluation, making it non-trivial to use for any other dataset; 3) despite our best efforts we could not get the model to converge to a good minima which would produce fluent text (besides word-by-word copying) when trained on the Shakespeare dataset.

Similarly, the datasets used in~\citet{sudhakar-etal-2019-transforming} modify semantic properties (sentiment, political slant etc.). On running their codebase on the Shakespeare dataset using the default hyperparameters, we achieved a poor performance of 53.1\% \accuracy, 55.2 \similarity~and 56.5\% \fluency, aggregating to a $J(\textsc{A,S,F})$ score of 18.4. Similarly on the Formality dataset, performance was poor with 41.7\% \accuracy, 67.8 \similarity~and 67.7\% \fluency, aggregating to $J(\textsc{A,S,F})$ score of 18.1. A qualitatively inspection showed very little abstraction and nearly word-by-word copying from the input (due to the delete \& generate nature of the approach), which explains the higher \similarity~score but lower \accuracy~score (just like \textsc{copy} baseline in \tableref{tab:prior-auto-eval}). Fluency was low despite GPT pretraining, perhaps due to the token deletion step in the algorithm.

\subsection{Details of our Dataset, \dataset}
\label{appendix:dataset}

We provide details of our sources, the sizes of individual style corpora and examples from our new benchmark dataset \dataset~in \tableref{tab:dataset-details}. We individually preprocessed each corpus to remove very short and long sentences, boilerplate text (common in Project Gutenberg articles) and section headings. We have added some representative examples from each style in \tableref{tab:dataset-details}. More representative examples (along with our entire dataset) will be provided in the project page \texttt{\href{http://style.cs.umass.edu}{http://style.cs.umass.edu}}.\\

\begin{figure}[t]
    \includegraphics[width=7cm]{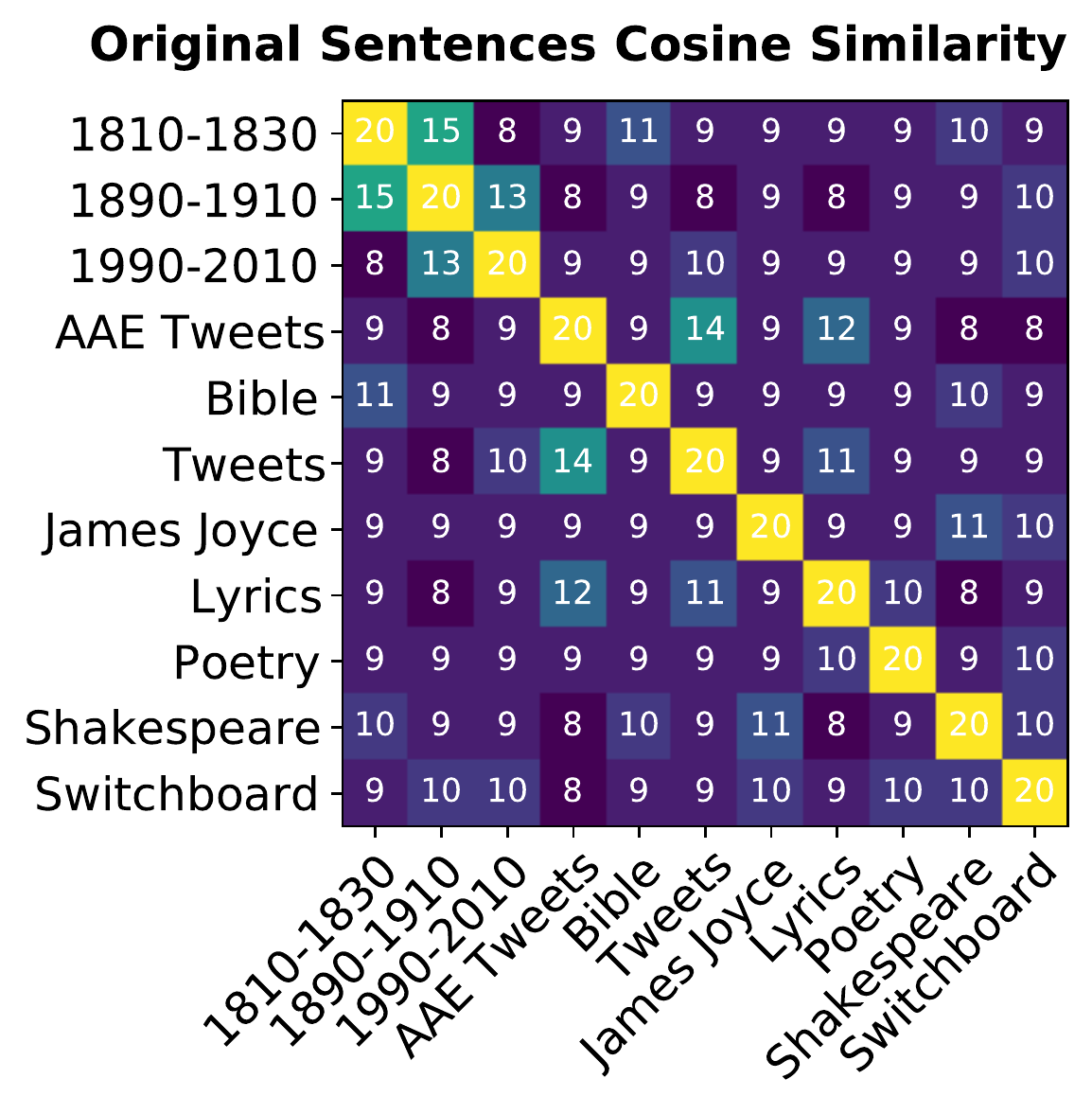}
    \caption{Cosine similarities between styles in \dataset~using the \texttt{[CLS]} vectors of the RoBERTa-large classifier (normalized to $[0, 20]$). The off-diagonal elements show intuitive domain similarities, such as (Lyrics, Poetry); (AAE, Tweets); (Joyce, Shakespeare) or among classes from the COHA corpus.
    }
    \label{fig:heatmaps-cosine}
\end{figure}

\noindent\textbf{Style Similarity:} In \figureref{fig:heatmaps-cosine} we plot the cosine similarity between styles using the averaged \texttt{[CLS]} vector of the trained RoBERTa-large classifier (inference over validation set). The off-diagonal elements show intuitive domain similarities, such as (Lyrics, Poetry); (AAE, Tweets); (Joyce, Shakespeare) or among classes from the Corpus of Historical American English.

\subsection{Diverse Paraphrasing on \dataset}
\label{appendix:paraphrase-our-dataset}

We compare the quality and diversity of the paraphrases generated by our diverse and non-diverse paraphrasers on our dataset \dataset~in \tableref{tab:our-dataset-pp-eval}. Note that this is the pseudo parallel training data for the inverse paraphrase model (described in \sectionref{sec:style-normalize} and \sectionref{sec:filtering-paraphrase}) and not the actual style transferred sentences. Overall, the diverse paraphraser achieves high diversity, with 51\% unigram change and 27\% word shuffling,\footnote{The ``unigram change'' and ``word shuffling'' refer to the unigram F1 word overlap and Kendall's $\tau_\text{B}$ scores.} compared to 28\% unigram and 6\% shuffling for non-diverse paraphraser, while maintaining good semantic similarity (\similarity = 72.5 vs 83.9 for non-diverse) even in complex stylistic settings.

\subsection{Style Transfer Performance on \dataset}
\label{appendix:performance-our-dataset}
We provide a detailed breakdown of performance in different styles of \dataset~in \tableref{tab:our-dataset-auto-eval}. For each of the 11 target styles, we style transferred 1,000 sentences from every other style and jointly evaluated the 10,000 generations. Some styles are more successfully transferred than others, such as Switchboard, Lyrics and James Joyce. While wearing the $p$ value for nucleus sampling, we notice a trend similar to the \textbf{Nucleus sampling trades off \accuracy~for \similarity} experiment in \sectionref{sec:ablation}. Increasing the $p$ value improves \accuracy~at the cost of \similarity. However unlike the Shakespeare and Formality dataset, we find $p=0.6$ the optimal value for the best \accuracy-\similarity~tradeoff.

Note that Fluency scores on this dataset could be misleading since even the original sentences from some styles are often classified as disfluent (Orig. \fluency). Qualitatively, this seems to happen for styles with rich lexical and syntactic diversity (like Romantic Poetry, James Joyce). These styles tend to be out-of-distribution for the fluency classifier trained on the CoLA dataset~\citep{warstadt2019neural}.

\subsection{A Survey of Evaluation Methods}
\label{appendix:eval-survey}

We present a detailed breakdown of evaluation metrics used in prior work in \tableref{tab:prior-eval-metric} and the implementations of the metrics in \tableref{tab:prior-eval-metric-implement}. Notably, only 3 out of 23 prior works use an absolute sentence-level aggregation evaluation. Other works either perform ``overall A/B'' testing, flawed corpus-level aggregation or don't perform any aggregation at all. Note that while ``overall A/B'' testing cannot be gamed like corpus-aggregation, it has a few issues --- (1) it is a \emph{relative} evaluation and does not provided an \emph{absolute} performance score for future reference; (2) ``A/B'' testing requires human evaluation, which is expensive and noisy; (3) evaluating overall performance will require human annotators to be familiar with the styles and style transfer task setup; (4)~\citet{kahneman2011thinking} has shown that asking humans to give a single number for ``overall score'' is biased when compared to an aggregation of \emph{independent} scores on different metrics. Luckily, the sentence-level aggregation in~\citet{li-etal-2018-delete} does the latter and is the closest equivalent to our proposed $J(\cdot)$ metric.

\subsection{Details on Human Evaluation}
\label{appendix:human-evaluation}

We conduct experiments of Amazon Mechanical Turk, annotating the paraphrase similarity of 150 sentences with 3 annotators each. We report the label chosen by two or more annotators, and collect additional annotations in the case of total disagreement. We pay workers 5 cents per sentence pair (\$10-15 / hr). We only hire workers from USA, UK and Australia with a  95\% or higher approval rating and at least 1000 approved HITs. Sentences where the input was exactly copied (after lower-casing and removing punctuation) are automatically assigned the option \textbf{2} paraphrase and grammatical. Even though these sentences are clearly not style transferred, we expect them to be penalized in $J$(\accuracy,\similarity,\fluency) by poor \accuracy. We found that every experiment had a Fleiss kappa~\citep{fleiss1971measuring} of at least 0.13 and up to 0.45 (slight to moderate agreement according to~\citep{landis1977measurement}). A qualitative inspection showed that crowdworkers found it easier to judge sentence pairs in the Formality dataset than Shakespeare, presumably due to greater familiarity with modern English. We also note that crowdworkers had higher agreement for sentences which were clearly not paraphrases (like the UNMT / DLSM generations on the Formality dataset).\\

\noindent\textbf{Calculating Metrics in \tableref{tab:prior-human-eval}:} To calculate \similarity, we count the percentage of sentences which humans assigned a label \textbf{1} (ungrammatical paraphrase) or \textbf{2} (grammatical paraphrase). This is used as a binary value to calculate $J$(\accuracy, \similarity). To calculate $J$(\accuracy, \similarity, \fluency), we count sentences which are correctly classified as well as humans assigned a label of \textbf{2} (grammatical paraphrase). We cannot calculate \fluency~alone using the popular 3-way evaluation, since the fluent sentences which are not paraphrases are not recorded.

\subsection{More Example Generations}
\label{appendix:example-generations}

More examples are provided in \tableref{tab:generations-2}. All of our style transferred outputs on \dataset~will be available in the project page of this work, \texttt{\href{http://style.cs.umass.edu}{http://style.cs.umass.edu}}.

\subsection{More Related Work}
\label{appendix:more_related}

Our inverse paraphrase model is a style-\textbf{controlled text generator} which automatically learns lexical and syntactic properties prevalent in the style's corpus. Explicit syntactically-controlled text generation has been studied previously using labels such as constituency parse templates~\citep{iyyer-etal-2018-adversarial, akoury-etal-2019-syntactically} or learned discrete latent templates~\citep{wiseman-etal-2018-learning}. 
Syntax can also be controlled using an exemplar sentence~\citep{chen-etal-2019-controllable,guu2018generating,peng-etal-2019-text}. While style transfer requires the underlying content to be provided as input, another direction explores attribute-controlled \emph{unconditional} text generation~\citep{dathathri2020plug,keskar2019ctrl,zeng2020style,ziegler2019fine}.\\

\noindent\textbf{Diversity in text generation} is often encouraged during inference time via heuristic modifications to beam search~\citep{li2016simple, vijayakumar2018diverse}, nucleus sampling~\citep{holtzman2020curious} or submodular optimization~\citep{kumar-etal-2019-submodular}; in contrast, we simply filter our training data to increase diversity. Other algorithms learn to condition generation on latent variables during training~\citep{bowman2016generating}, which are sampled from at inference time to encourage diversity~\citep{jain2017creativity, gupta2018deep, park2019paraphrase}. Relatedly,~\citet{goyal2020neural} promote syntactic diversity of paraphrases by conditioning over possible syntactic rearrangements of the input.

\begin{table*}[t]
\footnotesize
\begin{center}
\begin{tabular}{ lp{5.8cm}p{6.8cm} } 
\toprule
Direction & Input & Output \\
\midrule
1810. $\rightarrow$ Swit. & She forgot her gold pieces lying on the table in front of her. & and uh she forgot the gold pieces on the desk \\
1810. $\rightarrow$ AAE & This story is not well devised, my lord. & Lord this story ain't no good \\
Poetry $\rightarrow$ Swit. & Why should our young Endymion pine away! & i don't know why we should have such a letdown for our young Endymion \\
Tweet $\rightarrow$ Shak. & This is an insult to the Munster's. & Tis an affront to the Munster. \\
Swit. $\rightarrow$ 1810. & and my lawn is very green and my flowers are still blooming & I have plenty of green turf and flowers here. \\
1890. $\rightarrow$ AAE & I want no pursuit when I do my work. & I dont wanna chase when im at work \\
1990. $\rightarrow$ Joyce & ``You're scaring me,'' Clara says . & 'Tis scare me, says Clara. \\
Tweet $\rightarrow$ Bible & Can we build AI without losing control over it? & Can a man build a computer without losing his power? \\
AAE $\rightarrow$ 1810. & I Use To Think They Were Ugly. & I thought them odious--and hateful.\\
Bible $\rightarrow$ 1990. & hath thy soul lothed Zion? & Is Zion a place where you have your soul?\\
Bible $\rightarrow$ Tweet & This persuasion cometh not of him that calleth you. & Its not him who comes from. Its who you call \\
Swit. $\rightarrow$ AAE & um yeah just a second & yea just a lil bit \\
Joyce $\rightarrow$ Shak. & Smell of burn. & This is the smoke of a burning. \\
1990. $\rightarrow$ Joyce & The Second World War, son, " he answered. & SECOND WORLD WAR, son, says he. \\
1990. $\rightarrow$ AAE & no, I implore you not to consider this option. & Naw I asked u not to think bout that option \\
Tweet $\rightarrow$ Joyce & i wanted to sleep but now i cant & sleep is my preferred repose. \\
1990. $\rightarrow$ 1810. & She ninety-three and she can't hear too good. & She is 93 years of age and cannot hear well yet. \\
AAE $\rightarrow$ Poetry & y'all play? we spray & And weel weel, we'll spray a' that. \\
Joyce $\rightarrow$ Poetry & Now, don’t tell a big fib! & And say not so big a lie! \\
1810. $\rightarrow$ Bible & without first veiling that, I could not sin. & And I could not sin without veil.\\
1890. $\rightarrow$ Shak. & When did we ever find Gaston de Marignac objectionable? & When was ’t that we found Gaston de Marignac’s foul faults? \\
Shak. $\rightarrow$ 1990. & O, much I fear some ill unlucky thing. & I was afraid of a lot of misadventures. \\
Shak. $\rightarrow$ 1810. & By Jesu, a very good blade! & The very knife of heaven, Jesu! \\
Lyrics $\rightarrow$ 1890 & For a thing to explode & Oh, oh, oh, for something to blow up! \\
Lyrics $\rightarrow$ Bible & You are the sun that lights my world & Thou art the sun that lighteth mine earth: thou art my righteousness.\\
Lyrics $\rightarrow$ Joyce & But now it shouldn't have to be this way & But oughtn’t to be like that. \\
Swit. $\rightarrow$ 1810. & oh that poor child & (sighing) the poor dear little thing. \\
Swit. $\rightarrow$ Lyrics & you know Karen i wonder if we are recording & Karen, I wonder if we're tapein' \\
Tweet $\rightarrow$ Bible & Ok will get back to you ASAP! & And he said, All right; I will return.\\
Tweet $\rightarrow$ Shak. & Movies are always funnier in the cinema. & It always is more humorous in the theatre. \\
Lyrics $\rightarrow$ Poetry & You know I still fear the bomb & I was yet a-fear'd at the bomb, \\
Joyce $\rightarrow$ 1890. & Just missed that by a hair. & It was a hair-spitting thing I missed. \\
Joyce $\rightarrow$ 1810. & Me seemeth a dragon man. & Hark ye, I'm a dragon man, Me. \\
Poetry $\rightarrow$ Bible & Of deities or mortals, or of both, & Of the gods, and of the people, or of the both of them together. \\
Poetry $\rightarrow$ Lyrics & Makes countless thousands mourn! & Gonna be a lot of people mourning! \\
Poetry $\rightarrow$ Bible & My helpless lambs, I trust them wi' him. & Surely he will feed my lambs in safety. \\
1890. $\rightarrow$ Poetry & Stark was sentenced to ten years' imprisonment, Gibbon to five. & Stark was sentence'd ten yeares, Gibbon five; \\
1890. $\rightarrow$ Swit. & I want no pursuit when I do my work. & i don't want to chase after when i'm at work \\
Tweet $\rightarrow$ Bible & The way this is set up has me dying & This death cometh because of this. \\
Tweet $\rightarrow$ Swit. & Yay! Ill order them then! I dont have patience to wait on them! & now i'm gonna order those \\
Lyrics $\rightarrow$ Joyce & And all the guns is in Kesha's name & All your gunsees belong to Kesha. \\
AAE $\rightarrow$ 1810. & boredom gonna be the death of me one day & I shall die, one day, of the insupportable want of amusement. \\
AAE $\rightarrow$ 1890. & That's just what I needed to see.... Thank Ya Lord & Thank you, Lord; that is just what I was expecting. \\
AAE $\rightarrow$ Swit. & okay ii will see you later & yeah see you later
bye\\
Poetry $\rightarrow$ Tweet & Fam'd heroes! had their royal home: & royal bloods heroes:\\
Tweet $\rightarrow$ Bible & Check out this new painting that I uploaded to! & Look upon my new picture that I have set before thee! \\
Swit. $\rightarrow$ Shak. & so uh what do you wear to work & And what dost thou wear for thy work? \\
Tweet $\rightarrow$ Poetry & Now I gotta delete it & O now, must I part? And can I now erase \\
Tweet $\rightarrow$ 1810. & \#India is now producing the worlds cheapest solar power \#energy &  Now is India's solar power cheapest of all the world. \\
Poetry $\rightarrow$ Joyce & Away, away, or I shall dearly rue & O offside, away, or do I am rather sad.\\
Tweet $\rightarrow$ Swit. & Oh shit ima be a senior & so uh i got to the senior level of the business \\
\bottomrule
\end{tabular}
\end{center}
\caption{More example outputs from our model \model~trained on our dataset \dataset. Our project page will provide all 110k style transferred outputs generated by \model~on \dataset.}
\label{tab:generations-2}
\end{table*}

\begin{table*}[h!]
\begin{center}
\begin{tabular}{ lcccccccccc } 
 \toprule
Paper & \multicolumn{5}{c}{Automatic} & \multicolumn{5}{c}{Human} \\ 
\cmidrule(l){2-6} \cmidrule(l){7-11}
& \accuracy & \similarity & \fluency & \textsc{ca} & \textsc{sa} & \accuracy & \similarity & \fluency & \textsc{ca} & \textsc{sa} \\
\midrule
\citet{hu2017toward} & \checkmark \\
\citet{shen2017style} & \checkmark & & & & & \checkmark & & \checkmark & & A/B \\
\citet{shetty2018a4nt} & \checkmark & & & & & & A/B\\
\citet{fu2018style} & \checkmark & \checkmark & & & & & \checkmark\\
\citet{li-etal-2018-delete} & \checkmark & \checkmark & & & & \checkmark & \checkmark & \checkmark & & \checkmark \\
\citet{zhang2018style} & \checkmark & \checkmark & & & & \checkmark & \checkmark & \checkmark & & \checkmark\\
\citet{nogueira-dos-santos-etal-2018-fighting} & \checkmark & \checkmark & \checkmark \\
\citet{prabhumoye-etal-2018-style} & \checkmark & & & & & & A/B & \checkmark \\
\citet{xu-etal-2018-unpaired} & \checkmark & \checkmark & & \checkmark & & \checkmark & \checkmark & & \checkmark & \\
\citet{logeswaran2018content} & \checkmark & \checkmark & \checkmark & & & \checkmark & \checkmark & \checkmark \\
\citet{yang2018unsupervised} & \checkmark & \checkmark & \checkmark \\
\citet{subramanian2018multiple} & \checkmark & \checkmark & \checkmark & & & \checkmark & \checkmark & \checkmark & & A/B\\
\citet{luo2019dual} & \checkmark & \checkmark &  & \checkmark & & \checkmark & \checkmark & \checkmark & \checkmark & \checkmark \\
\citet{pang-gimpel-2019-unsupervised} & \checkmark & \checkmark & \checkmark & \checkmark & & A/B & A/B & A/B & & A/B\\
\citet{ma2019syntax-style} & \checkmark & \checkmark & \checkmark & & & \checkmark & \checkmark & \checkmark \\
\citet{dai-etal-2019-style} & \checkmark & \checkmark & \checkmark & & & A/B & A/B & A/B \\
\citet{sudhakar-etal-2019-transforming} & \checkmark & \checkmark & \checkmark & & & A/B & A/B & A/B & & A/B\\
\citet{mir-etal-2019-evaluating} & \checkmark & \checkmark & \checkmark & & & \checkmark & \checkmark & \checkmark \\
\citet{grondahl2019effective} & \checkmark & \checkmark & & & & & \checkmark & \\
\citet{tikhonov2019style} & \checkmark & \checkmark \\ 
\citet{syed2020adapting} & \checkmark & \checkmark \\
\citet{madaan2020politeness} & \checkmark & \checkmark & & & & \checkmark & \checkmark & \checkmark \\
\citet{he2020probabilistic} & \checkmark & \checkmark & \checkmark \\
\midrule
Ours & \checkmark & \checkmark & \checkmark & \checkmark & \checkmark & & \checkmark & \checkmark &  & \checkmark \\
\bottomrule
\end{tabular}
\end{center}
\caption{Survey of evaluation methods used in 23 prior papers. We check whether prior work evaluate their algorithm on transfer accuracy (\accuracy), semantic similarity (\similarity), fluency (\fluency), corpus-level aggregation (\textsc{ca}) and sentence-level aggregation (\textsc{sa}). We use the ``A/B'' to denote relative comparisons via A/B testing between generations from the baseline and the proposed system, rather than absolute performance numbers. Specific implementations of the metrics have been provided in \tableref{tab:prior-eval-metric-implement}. We do not include~\citet{pang-2019-towards} since it's a survey of existing evaluation methods.}
\label{tab:prior-eval-metric}
\end{table*}

\begin{table*}[h!]
\small
\begin{center}
\begin{tabular}{ lcccccc } 
 \toprule
Paper & \multicolumn{3}{c}{Automatic} & \multicolumn{3}{c}{Human} \\ 
\cmidrule(l){2-4} \cmidrule(l){5-7}
& \accuracy & \similarity & \fluency & \accuracy & \similarity & \fluency \\
\midrule
\citet{hu2017toward} & L-CNN \\
\citet{shen2017style} & CNN & & & Likert-4 &  & Likert-4\\
\citet{shetty2018a4nt} & RNN/CNN & METEOR & & & A/B & \\
\citet{fu2018style} & LSTM & GloVE & & & Likert-3 &\\
\citet{li-etal-2018-delete} & LSTM & BLEU & & Likert-5 & Likert-5 & Likert-5 \\
\citet{zhang2018style} & GRU & BLEU & & Likert-5 & Likert-5 & Likert-5\\
\citet{nogueira-dos-santos-etal-2018-fighting} & SVM & GloVE & PPL \\
\citet{prabhumoye-etal-2018-style} & CNN & & & & A/B & Likert-4 \\
\citet{xu-etal-2018-unpaired} & CNN & BLEU & & Likert-10 & Likert-10 & \\
\citet{logeswaran2018content} & CNN & BLEU & PPL & Likert-5 & Likert-5 & Likert-5 \\
\citet{yang2018unsupervised} & CNN & BLEU & PPL \\
\citet{subramanian2018multiple} & fastText & BLEU & PPL & Binary & Likert-5 & Likert-5\\
\citet{luo2019dual} & CNN & BLEU & & Likert-5 & Likert-5 & Likert-5 \\
\citet{pang-gimpel-2019-unsupervised} & CNN & GloVE & PPL & A/B & A/B & A/B \\
\citet{ma2019syntax-style} & CNN & BLEU & PPL & Likert-5 & Likert-5 & Likert-5 \\
\citet{dai-etal-2019-style} & fastText & BLEU & PPL & A/B & A/B & A/B\\
\citet{sudhakar-etal-2019-transforming} & fastText & GLEU & PPL & A/B & A/B & A/B\\
\citet{mir-etal-2019-evaluating} & EMD & GloVE* & Classify & Likert-5* & Likert-5* & Binary* \\
\citet{grondahl2019effective} &  LSTM/CNN & METEOR\\
\citet{tikhonov2019style} & CNN & BLEU \\ 
\citet{syed2020adapting} & FineGrain & BLEU \\
\citet{madaan2020politeness} & AWD-LSTM & METEOR & & Likert-5 & Likert-5 & Likert-5\\
\citet{he2020probabilistic} & CNN & BLEU & PPL &  \\
\midrule
Ours &  RoBERTa-L & SIM-PP & Classify & & Binary & Binary \\
\bottomrule
\end{tabular}
\end{center}
\caption{Survey of implementations of evaluation metrics to measure Accuracy (\accuracy), Similarity (\similarity) and Fluency (\fluency) used in 23 prior papers. For a cleaner version of this table with aggregation information, see \tableref{tab:prior-eval-metric}. The * marks in~\citet{mir-etal-2019-evaluating} denote a carefully designed unique implementation. We do not include~\citet{pang-2019-towards} since it's a survey of existing evaluation methods.}
\label{tab:prior-eval-metric-implement}
\end{table*}

\begin{table*}[t!]
\small
\begin{center}
\begin{tabular}{ lrrrrr|rrrrr } 
 \toprule
 Model & \multicolumn{5}{c|}{Formality} & \multicolumn{5}{c}{Shakespeare} \\
 & \accuracy & \similarity & \fluency & \textsc{gm}(\textsc{a,s,f}) & $J$(\textsc{a,s,f}) & \accuracy & \similarity & \fluency & \textsc{gm}(\textsc{a,s,f}) & $J$(\textsc{a,s,f}) \\
 \midrule
\textsc{copy} & 5.2 & 41.8 & 88.4 & 26.8 & 0.2 & 9.6 & 20.1 & 79.1 & 24.8 & 0.1 \\
\textsc{na\"ive} & 49.7 & 22.1 & 89.4 & 44.4 & 2.4 & 49.9 & 10.5 & 78.9 & 34.6 & 1.1\\
\textsc{ref} & 93.3 & 100 & 89.7 & 94.2 & 88.2 & 90.4 & 100 & 79.1 & 89.4 & 67.2 \\
\midrule
\textsc{unmt} & 78.5 & 15.1 & 52.5 & 39.7 & 11.7 & 70.5 & 7.9 & 49.6 & 30.2 & 1.7 \\
\textsc{dlsm} & 78.0 & 18.5 & 53.7 & 42.6 & 9.5 & 71.1 & 12.5 & 49.4 & 35.2 & 2.0 \\
\midrule
\model~($p = 0.0$) & 67.7 &  28.8 & 90.4 & 56.1 & 19.3 & 71.7 & 10.3 & 85.2 & 39.8 & 5.9 \\
\model~($p = 0.6$) & 70.7 & 25.3 & 88.5 & 54.1 & 17.2 & 75.7 & 8.8 & 82.7 & 38.1 & 5.4 \\
\model~($p = 0.9$) & 76.8 & 17.0 & 77.4 & 46.6 & 12.2 & 79.8 & 6.1 & 71.7 & 32.7 & 3.4 \\
\bottomrule
\end{tabular}
\end{center}
\caption{A table equivalent to \tableref{tab:prior-auto-eval} but using BLEU scores for \similarity~instead of the paraphrase similarity model from \citet{wieting-etal-2019-beyond}. The Formality dataset had 4 available reference sentences whereas the Shakespeare dataset had only 1 available reference sentence. Our system \model~significantly beats prior work (\textsc{unmt}, \textsc{dlsm}) on $J(\cdot)$ metrics even with BLEU scores.}
\label{tab:prior-auto-eval-bleu}
\end{table*}

\begin{table*}[t!]
\small
\begin{center}
\begin{tabular}{ lrrrrrrrrrr } 
 \toprule
Model & \accuracy~(\textsc{a}) & \multicolumn{2}{c}{\similarity~(\textsc{s})} & \fluency~(\textsc{f}) & \multicolumn{2}{c}{$J$(\textsc{a,s})} & \multicolumn{2}{c}{$J$(\textsc{a,s,f})}  \\ 
\cmidrule(l){3-4} \cmidrule(l){6-7} \cmidrule(l){8-9}
& & \textsc{bl} & \textsc{pp} &  & \textsc{bl} & \textsc{pp} & \textsc{bl} & \textsc{pp}  \\ 
\midrule
\textsc{copy} & 8.0 & 32.6 & 80.9 & 90.1 & 0.4 & 7.1 & 0.3& 6.4\\
\textsc{ref} & 87.8 & 100 & 100 & 90.1 & 91.1 & 87.8 & 83.5 & 78.9 \\
\textsc{na\"ive} & 67.9 & 10.7 & 32.0 & 91.5 & 1.7 & 9.3 & 1.5 & 8.5 \\
 \midrule
 BT \citep{prabhumoye-etal-2018-style} & 47.4 & 1.3 & 21.1 & 8.0 & 0.7 & 11.4 & 0.0 & 1.3 \\
 MultiDec \citep{fu2018style} & 26.0 & 12.0 & 36.9 & 15.1 & 1.4 & 8.9 & 0.0 & 1.5 \\
 Del. \citep{li-etal-2018-delete} & 24.2 & 30.1 & 53.5 & 20.8 & 3.1 & 10.2 & 0.0 & 1.6\\
Unpaired \citep{xu-etal-2018-unpaired} & 53.9 & 1.6 & 16.3 & 34.9 & 0.4 & 10.9 & 0.0 & 2.2 \\
DelRetri. \citep{li-etal-2018-delete} & 52.8 & 21.9 & 47.6 & 16.3 & 11.9 & 23.4 & 0.2 & 4.2\\
CrossAlign. \citep{shen2017style} & 59.0 & 3.3 & 25.0 & 31.7 & 2.0 & 14.9 & 0.3 & 5.2\\
Retri. \citep{li-etal-2018-delete} & 90.0 & 0.5 & 9.0 & 62.1 & 0.5 & 8.3 & 0.3 & 5.5\\
Templ. \citep{li-etal-2018-delete} & 37.1 & 36.4 & 67.8 & 32.3 & 11.9 & 23.7 & 1.3 & 7.8 \\
DualRL \citep{luo2019dual} & 51.8 & 45.0 & 65.1 & 59.0 & 14.6 & 29.9 & 8.1 & 21.7\\
UNMT \citep{zhang2018style} & 64.5 & 34.4 & 64.8 & 45.9 & 28.2 & 41.2 & 14.7 & 22.1 \\
\midrule 
\model~($p = 0.0$)* & 57.7 & 31.1 & 69.7 & 93.8 & 19.5 & 40.8 & \textbf{18.3} & 38.7 \\
\model~($p = 0.6$)* & 63.4 & 26.5 & 66.7 & 91.4 & 18.3 & \textbf{43.0} & 17.1 & \textbf{40.0} \\
\model~($p = 0.9$)* & 70.3 & 17.3 & 59.0 & 81.4 & 13.6 & 41.6 & 11.8 & 34.3 \\
\bottomrule
\end{tabular}
\end{center}
\caption{More comparisons against prior work on the Formality dataset~\citep{rao-tetreault-2018-dear} using the outputs provided in the publicly available codebase of~\citet{luo2019dual} using both BLEU score (\textsc{bl}) and paraphrase similarity (\textsc{pp}). This model uses the Family \& Relationships split of the Formality dataset whereas~\citep{he2020probabilistic} used the Entertainment \& Music split. Hence, we have retrained our RoBERTa-large classifiers to reflect the new distribution. \textbf{*Note}: While our system significantly outperforms prior work, we re-use the formality system used in \tableref{tab:prior-auto-eval} and \tableref{tab:prior-human-eval} for these results, which was trained on Entertainment \& Music (consistent with \citet{he2020probabilistic}). There could be a training dataset mismatch between our model and the models from~\citet{luo2019dual}, since the Formality dataset has two domains. This is not clarified in~\citet{luo2019dual} to the best of our knowledge.}
\label{tab:prior-auto-eval-luo}
\end{table*}

\begin{table*}[h!]
\small
\begin{center}
\begin{tabular}{ p{2cm}rrrp{4cm}p{4cm} }
\toprule
Style & Train & Dev & Test & Source & Examples\\

\midrule
Shakespeare & 24,852 & 1,313 & 1,293 & Shakespeare split of \citet{xu-etal-2012-paraphrasing}. & 1. \textit{Why, Romeo, art thou mad?} \newline 2. \textit{I beseech you, follow straight.}\\
\midrule
English Tweets & 5,164,874 & 39,662 & 39,690 & A random sample of English tweets collected on 8th-9th~July, 2019 using Twitter APIs. & 1. \textit{Lol figures why I dont wanna talk to anyone rn} \newline 2. \textit{omg no problem i felt bad holding it! i love youuuu} \\
\midrule
Bible & 31,404 & 1,714 & 1,714 & The English Bible collected from Project Gutenberg~\citep{hart1992history} (\href{http://www.gutenberg.org/cache/epub/10/pg10.txt}{link}). & 1. \textit{Jesus saith unto her, Woman, what have I to do with thee?} \newline 2. \textit{Wherefore it is lawful to do well on the sabbath days.}\\
\midrule
Romantic Poetry & 26,880 & 1,464 & 1,470 & The Romantic section of the Poetry bookshelf on Project Gutenberg  (\href{https://www.gutenberg.org/wiki/Poetry_(Bookshelf)#Romantic}{link}). & 1. \textit{There in that forest did his great love cease;} \newline 2. \textit{But, oh! for Hogarth's magic pow'r!}\\
\midrule
Switchboard & 145,823 & 1,487 & 1,488 & Conversational speech transcripts (\href{https://www.isip.piconepress.com/projects/switchboard/}{link}) from the Switchboard speech recognition corpus~\citep{godfrey1992switchboard}. & 1. \textit{uh-huh well we're not all like that um} \newline 2. \textit{well yes i i well i- i don't think i have the time to really become a student in every article}\\
\midrule
AAE (African American English) Tweets & 717,634 & 7,316 & 7,315 & Using the geo-located tweet corpus collected by~\citet{blodgett-etal-2016-demographic}. & 1. \textit{ay yall everything good we did dat...} \newline 2. \textit{I know data right, it don't get more real than that.}\\
\midrule
James Joyce & 37,082 & 2,054 & 2,043 & Two novels (Ulysses, Finnegans) of James Joyce from Project Gutenberg (\href{http://www.gutenberg.org/ebooks/4300}{link}) and the Internet Archive (\href{https://archive.org/stream/finneganswake00joycuoft/finneganswake00joycuoft_djvu.txt}{link}). & 1. \textit{At last she spotted a weeny weeshy one miles away.} \newline 2. \textit{chees of all chades at the same time as he wags an antomine art of being rude like the boor.}\\
\midrule
Lyrics & 4,588,522 & 252,368 & 252,397 & Music lyrics dataset from MetroLyrics, used in a Kaggle competition (\href{https://www.kaggle.com/gyani95/380000-lyrics-from-metrolyrics}{link}). & 1. \textit{I gotta get my mind off you,} \newline 2. \textit{This is it, we are, baby, we are one of a kind}\\
\midrule
1810-1830 historical English & 205,286 & 5,340 & 5,338  & 1810-1830 in the Corpus of Historical American English~\citep{davies2012expanding} using fiction, non-fiction and magazine domains (\href{https://www.english-corpora.org/coha}{link}). & 1. \textit{The fulness of my fancy renders my eye vacant and inactive.} \newline 2. \textit{What then do you come hither for at such an hour?}\\
\midrule
1890-1910 historical English & 1,210,687 & 32,024 & 32,018  & 1890-1910 in the Corpus of Historical American English using fiction, non-fiction and magazine domains (\href{https://www.english-corpora.org/coha}{link}). & 1. \textit{Nor shall I reveal the name of my friend; I do not wish to expose him to a torrent of abuse.} \newline 2. \textit{You know olive oil don't give the brightest illumination.}\\
\midrule
1990-2010 historical English & 1,865,687 & 48,985 & 48,982 & 1990-2010 in the Corpus of Historical American English using fiction, non-fiction and magazine domains (\href{https://www.english-corpora.org/coha}{link}). & 1. \textit{They were, in fact, tears of genuine relief.} \newline 2. \textit{I don't know why, but I sensed there was something wrong.}\\
\midrule
\textbf{Total} & 14,018,731 & 393,727 & 393,748 & \\
\bottomrule
\end{tabular}
\end{center}
\caption{Details of our new benchmark dataset \dataset~along with representative examples. Our dataset contains eleven lexically and syntactically diverse styles and has a total of nearly 15M sentences, an order of magnitude larger than previous datasets. We will provide more representative examples along with our entire dataset in the project page \texttt{\href{http://style.cs.umass.edu}{http://style.cs.umass.edu}}.}
\label{tab:dataset-details}
\end{table*}

\begin{table*}[t!]
\small
\begin{center}
\begin{tabular}{ lrrrrrrrr } 
 \toprule
Split & Orig. \accuracy & Orig. \fluency  & Model & \accuracy~(\textsc{a}) & \similarity~(\textsc{s}) & \fluency~(\textsc{f}) & $J$(\textsc{a,s}) & $J$(\textsc{a,s,f}) \\ 
\midrule
AAE Tweets & 87.6 & 56.4 & Ours ($p = 0.0$) & 21.0 & 70.1 & 71.6 & 12.6 & 8.3\\
& && Ours ($p = 0.6$) &  32.5 & 65.7 & 63.5 & 18.3 & \textbf{10.2} \\
& && Ours ($p = 0.9$) & 46.1 & 57.8 & 45.9 & \textbf{23.6} & 9.8\\
\midrule
Bible & 98.3 & 87.5 & Ours ($p = 0.0$) & 48.0 & 58.4 & 81.2 & 24.7 & 20.9\\
& && Ours ($p = 0.6$) & 52.5 & 55.1 & 79.8 & \textbf{25.7} & \textbf{21.3} \\
& && Ours ($p = 0.9$) & 56.9 & 49.4 & 74.0 & 25.3 & 19.3 \\
\midrule
COHA 1810s-1820s & 83.0 & 89.1 & Ours ($p = 0.0$) & 25.9 & 66.5 & 84.5 & 16.4 & 13.7\\
& && Ours ($p = 0.6$) & 34.0 & 63.0 & 81.5 & 20.1 & 16.0 \\
& && Ours ($p = 0.9$) & 42.7 & 57.3 & 73.6 & \textbf{22.9} & \textbf{16.5} \\
\midrule
COHA 1890s-1900s & 76.5 & 91.2 & Ours ($p = 0.0$) & 36.1 & 68.9 & 86.7 & 23.7 & 21.2\\
& && Ours ($p = 0.6$) & 41.1 & 65.7 & 83.8 & \textbf{25.5} & \textbf{22.1} \\
& && Ours ($p = 0.9$) & 44.3 & 59.4 & 72.0 & 25.0 & 19.2 \\
\midrule
COHA 1990s-2000s & 86.9 & 96.8 & Ours ($p = 0.0$) & 40.4 & 69.0 & 87.7 & 26.6 & 24.4\\
& && Ours ($p = 0.6$) & 46.1 & 65.6 & 86.0 & \textbf{28.9} & \textbf{26.3} \\
& && Ours ($p = 0.9$) & 46.1 & 59.4 & 76.1 & 26.1 & 21.7\\
\midrule
English Tweets & 80.7 & 79.9 & Ours ($p = 0.0$) & 20.0 & 71.0 & 79.1 & 13.5 & 11.0\\
& && Ours ($p = 0.6$) & 28.9 & 67.5 & 72.2 & 18.1 & \textbf{13.7}\\
& && Ours ($p = 0.9$) & 40.8 & 60.0 & 55.5 & \textbf{22.7} & 13.4\\
\midrule
James Joyce & 87.1 & 48.2 & Ours ($p = 0.0$) & 43.0 & 69.6 & 79.8 & 28.7 & 22.0 \\
& && Ours ($p = 0.6$) & 52.2 & 63.7 & 62.8 & 32.0 & \textbf{29.6} \\
& && Ours ($p = 0.9$) & 63.6 & 54.8 & 40.5 & \textbf{33.5} & 11.3\\
\midrule
Lyrics & 88.7 & 78.9 & Ours ($p = 0.0$) & 51.9 & 71.6 & 79.4 & \textbf{35.6} & \textbf{29.0}\\
& && Ours ($p = 0.6$) & 53.4 & 68.6 & 71.4 & 34.8 & 26.0 \\
& && Ours ($p = 0.9$) & 53.3 & 62.1 & 51.9 & 31.4 & 18.1\\
\midrule
Romantic Poetry & 93.8 & 40.2 & Ours ($p = 0.0$) & 55.0 & 63.8 & 58.9 & 33.5 & \textbf{17.2}\\
& && Ours ($p = 0.6$) & 62.4 & 60.3 & 51.8 & 35.6 & 16.2 \\
& && Ours ($p = 0.9$) & 69.8 & 55.3 & 40.3 & \textbf{36.8} & 13.0 \\
\midrule
Shakespeare & 86.1 & 59.9 & Ours ($p = 0.0$) & 36.8 & 65.5 & 76.9 & 21.7 & 15.4 \\
& && Ours ($p = 0.6$) & 52.1 & 58.6 & 65.4 & 28.2 & \textbf{16.6} \\
& && Ours ($p = 0.9$) & 63.7 & 48.9 & 44.2 & \textbf{29.3} & 11.3 \\
\midrule
Switchboard & 99.7 & 63.1 & Ours ($p = 0.0$) & 62.9 & 67.4 & 77.0 & 40.8 & 32.0\\
& && Ours ($p = 0.6$) & 77.2 & 63.7 & 64.2 & \textbf{47.5} & \textbf{30.2} \\
& && Ours ($p = 0.9$) & 84.9 & 56.6 & 44.0 & 46.8 & 20.1 \\
\midrule
\textbf{Overall} & 88.0 & 71.9 & Ours ($p = 0.0$) & 40.1 & 67.4 & 78.4 & 25.3 & 19.6\\
& && Ours ($p = 0.6$) & 48.4 & 63.4 & 71.1 & 28.6 & \textbf{20.7}\\
& && Ours ($p = 0.9$) & 55.7 & 56.5 & 56.2 & \textbf{29.4} & 15.8 \\
\bottomrule
\end{tabular}
\end{center}
\caption{A detailed performance breakup when transferring \textbf{to} each style in \dataset~from the other 10 styles. We test three nucleus sampling~\citep{holtzman2020curious} strategies with our trained model by varying the $p$ value between 0.0 (greedy) and 1.0 (full sampling). For reference, the classification accuracy (Orig. \accuracy) and fluency (Orig. \fluency) of original sentences in the target style corpus are provided.}
\label{tab:our-dataset-auto-eval}
\end{table*}

\begin{table*}[t!]
\small
\begin{center}
\begin{tabular}{ lrrr|rrr } 
 \toprule
 & \multicolumn{3}{c|}{Diverse Paraphraser} &  \multicolumn{3}{c}{Non-Diverse Paraphraser}\\
Split & Similarity $(\uparrow)$ & Lexical $(\downarrow)$ & Syntactic $(\downarrow)$ & Similarity $(\uparrow)$ & Lexical $(\downarrow)$ & Syntactic $(\downarrow)$\\ 
\midrule
AAE Tweets & 65.1 & 44.7 & 0.43 & 74.3 & 66.4 & 0.82\\
Bible & 74.6 & 48.5 & 0.55 & 88.3 & 73.5 & 0.92 \\
COHA 1810s-1820s & 74.0 & 50.6 & 0.51 & 86.3 & 71.8 & 0.92 \\
COHA 1890s-1900s & 75.3 & 52.0 & 0.50 & 88.2 & 75.3 & 0.93 \\
COHA 1990s-2000s & 77.6 & 57.4 & 0.53 & 89.9 & 80.7 & 0.95 \\
English Tweets & 73.1 & 52.4 & 0.50 & 82.8 & 75.7 & 0.91 \\
James Joyce & 71.5 & 47.8 & 0.35 & 82.4 & 69.8 & 0.82 \\
Lyrics & 74.5 & 52.8 & 0.52 & 86.7 & 78.6 & 0.92\\
Romantic Poetry & 72.3 & 46.3 & 0.44 & 81.3 & 67.1 & 0.86 \\
Shakespeare & 67.9 & 38.7 & 0.23 & 81.4 & 63.4 & 0.75 \\
Switchboard & 71.6 & 50.1 & 0.55 & 81.1 & 72.4 & 0.90\\
\midrule
\textbf{Overall} & 72.5 & 49.2 & 0.46 & 83.9 & 72.3 & 0.88 \\
\bottomrule
\end{tabular}
\end{center}
\caption{A detailed style-wise breakup of the \textbf{diverse paraphrase quality} in \dataset~(the training data for the inverse paraphrase model, described in \sectionref{sec:style-normalize} and \sectionref{sec:filtering-paraphrase}). The ideal paraphraser should score lower on ``Lexical'' and ``Syntactic'' overlap and high on ``Similiarity''. Overall, our method achieves high diversity (51\% unigram change and 27\% word shuffling, compared to 28\% unigram and 6\% shuffling for non-diverse), while maintaining good semantic similarity (\similarity = 72.5 vs 83.9 for non-diverse) even in complex stylistic settings. We measure lexical overlap in terms of unigram F1 overlap using the evaluation scripts from~\citet{rajpurkar-etal-2016-squad}. Syntactic overlap is measured using Kendall's $\tau_\text{B}$~\citep{kendall1938new} of shared vocabulary. A $\tau_\text{B} = 1.0$ indicates no shuffling whereas a value of $\tau_\text{B} = -1.0$ indicates 100\% shuffling (complete reversal). Finally, the \textsc{sim} model from~\citet{wieting-etal-2019-beyond} is used for measuring similarity.}
\label{tab:our-dataset-pp-eval}
\end{table*}

\end{document}